\documentclass[11pt,a4paper]{article}

\usepackage[T1]{fontenc}
\usepackage[utf8]{inputenc}
\usepackage{mathptmx} % Times-like; swap for newtxtext,newtxmath if available
\usepackage[margin=1in]{geometry}
\usepackage{graphicx}
\usepackage{booktabs}
\usepackage{amsmath,amssymb}
\usepackage{siunitx}
\usepackage{caption}
\usepackage[dvipsnames]{xcolor}
\usepackage[colorlinks=true,linkcolor=BrickRed,citecolor=BrickRed,urlcolor=NavyBlue]{hyperref}
\usepackage{cite}
\usepackage{microtype}
\usepackage{tikz}
\usetikzlibrary{positioning,arrows.meta,fit,backgrounds,calc}
\usepackage{authblk}

\let\oldthebibliography\thebibliography
\renewcommand{\thebibliography}[1]{%
 \oldthebibliography{#1}%
 \setlength{\itemsep}{0pt}\setlength{\parsep}{0pt}\setlength{\parskip}{0pt}}

\newif\ifdraft\draftfalse

\title{\bfseries Nothing to Disagree About:\\
Ensemble Disagreement as an Uncertainty Estimate\\
in Frozen-Feature World Models}

\author[1]{Faraz Khan}
\author[1]{Harun \v{S}iljak}
\affil[1]{\small School of Engineering, Trinity College Dublin\protect\\
\texttt{faraz002khan@gmail.com}\quad\texttt{harun.siljak@tcd.ie}}
\date{\today}

\begin{document}
\maketitle

\begin{abstract}
\noindent
A robot acting through a learned world model has to decide when to stop
trusting it, and that decision matters most in the case the model was never
shown: a floor that has turned slippery, a heavier object, a worn gripper. In
world models built on frozen self-supervised encoders, a common substrate for
robot planning, the decision has no native signal to rest on: these models
predict point estimates. The signal the architecture leaves available as an
uncertainty estimate is the disagreement of an ensemble of latent dynamics
predictors, carried over from state-space and reconstructive world models,
where it penalises imagined trajectories, shortens horizons or vetoes unsafe
actions. Such models never decode back to pixels, so both the estimate and the
error it should anticipate live in a learned space, and whether one tracks the
other in the physical world has not, to our knowledge, been tested.
We test it on DINO-WM and Push-T, probing latents to simulator state so that
disagreement is scored in physical units. It proves a fair estimate of
difficulty and a poor detector of novelty. One step ahead in distribution it
ranks physical error well ($\rho = 0.56$), but over an imagined rollout, the mode a planner actually runs in, the ranking
decays to near chance.
And when the world itself changes it moves the wrong way: a slippery floor more
than doubles physical error while disagreement \emph{falls} by a quarter, and a
lighting change that leaves the physics untouched raises it more than any
change in friction does. It follows how much the block is moving; a change in the dynamics leaves it unmoved. Changing what it sees shows why: the
predictor is handed the pusher's own velocity and does not read how the block
responded, where a change in dynamics shows up.
 For deployment, the signal is quietest exactly when the
world has shifted under the robot, so a planner that replans on real observations proceeds confidently into the conditions the gate exists for. For what to build next,
members that do not represent a change in dynamics have nothing to disagree
about however they are initialised, so a more diverse ensemble would not repair
this. A remedy would have to be architectural, in predictors that read how
objects respond to what they did, alongside trust grounded in quantities a robot can measure
rather than in a model's account of its own confidence.
\end{abstract}

% ============================================================
\section{Introduction}\label{sec:intro}
% ============================================================

A robot that plans through a learned model of its environment is only as safe
as its sense of when that model has stopped applying. The situations where this
matters are the ones no training set covers: a floor that has turned slippery,
a heavier object than before, a gripper that has worn. A planner that notices
has options: replan, slow down, ask for help. A planner that does not will
commit to a trajectory its model cannot predict, and will do so confidently.

That sense of when to stop trusting a model is not an accessory to planning. It
is load-bearing in the methods built to make a learned model safe to act on:
MOReL \cite{morel} ends a rollout in a penalty state once the model is judged
to be outside what it knows, MOPO \cite{mopo} subtracts an uncertainty estimate
from the reward, LOMPO \cite{lompo} does the same in a learned latent space,
Plan2Explore \cite{plan2explore} inverts the sign and steers towards
uncertainty to explore, and a latent safety filter \cite{safetyfilter} vetoes
actions whose outcomes it cannot certify as safe. Each of these converts an
uncertainty estimate into a decision, and each is only as good as the estimate
it is handed.

In one current family of robot world models that estimate is unusually hard to
come by. These models place a small action-conditioned predictor on top of a
\emph{frozen} self-supervised visual encoder \cite{dinowm,dinoworld,vjepa2}: an
image is encoded to a \emph{latent} state, the predictor maps a latent and an
action to the next latent, and a sampling-based planner searches for the action
sequence whose predicted latent best matches an encoded goal. Training never
decodes back to pixels, since the target is the encoding of the observation
that followed, which keeps representation learning separate from
dynamics learning and is cheap enough to train from offline video alone, with
no rewards and no demonstrations \cite{dinowm}, which is much of why the approach
is being adopted.
The cost is that every quantity such a system computes about itself, its error
and its confidence alike, is expressed in a learned space rather than in the
world it acts on. A small number in that space may or may not mean the block is
where the model expected.

These predictors emit a single point, so uncertainty has to be added after the
fact, and the architecture leaves few places to add it. The predictor is deterministic by construction, a limitation
DINO-world \cite{dinoworld} names explicitly, so there is no predictive
distribution whose variance could be read, as there is in the probabilistic
models MOPO \cite{mopo} and LOMPO \cite{lompo} are built on. Reading a latent
out into pixels or into state is possible, and the original implementation provides decoder heads for
it, but a readout returns a prediction; turning that into an error takes the observation that followed. That is the residual Cao and Bansal
\cite{worldmodelslie} build on, and it is informative, but it arrives only once
the step has been taken, while a planner needs its estimate with the rollout
still imagined. Distance to the training distribution in feature space is cheap
and available in advance, and it is the basis of a well-established line of
out-of-distribution detection: a Mahalanobis distance to class-conditional
Gaussians fitted to the penultimate features of a trained network
\cite{mahalanobis}, or a density estimator fitted to those features directly
\cite{ddu}. Both score a single input by its appearance, and a change in
friction leaves the image unchanged.
What remains is the construction the
disagreement literature already supplies: train an \emph{ensemble} of
predictors differing only in initialisation and data order, and read the spread
of their predictions, their \emph{disagreement}, as the model's epistemic
uncertainty \cite{plan2explore,morel,mopo,lompo,safetyfilter}. The assumption has two steps, and in the work we examine only the first is
made explicit. Spread is
wide where the training data was thin, so MOReL's detector is a
support test over the training distribution \cite{morel} and why methods that
make the same reading explicit estimate density in feature space directly
\cite{ddu}. The unstated second step is that novelty in the world shows up as
novelty in the input. Our result is that in this class of models those two
dissociate.

Two recent results call that assumption into question without resolving it. Berger et
al.\ \cite{biaseddreams} find that in RSSM-style recurrent world models,
disagreement captures local epistemic uncertainty but does not track the
compounding error of a long latent rollout: an attractor pulls rollouts toward
well-supported latent regions, where disagreement shrinks while the rollout
drifts further from the true dynamics, and returns are overestimated wherever
those regions coincide with high-reward behaviour. They put the transfer
itself in question, arguing that uncertainty estimates do not carry over from
proprioceptive dynamics models to latent ones as the literature assumes.
Vakalis \cite{intervention} audits several substrates and includes the one
nearest ours, five independently initialised action-conditioned predictors over
a shared frozen DINOv2 backbone, reporting that their disagreement ranks future
\emph{feature} error under task-policy support ($\rho = 0.678$) and is
inconclusive under environment-random support ($\rho = 0.027$, interval
spanning zero).

Feature error is the natural target if the question is whether an ensemble
predicts its own loss, but it is not the quantity a planner needs. Epistemic uncertainty is defined
relative to what a model can in principle resolve from its own inputs
\cite{hullermeier}, so a quantity it cannot read behaves towards it like
irreducible noise however reducible that quantity is in the world. A signal
that tracks feature error faithfully can therefore stay silent about a physical
change the features never delivered to the predictor. Whether the two part
company in practice we are not aware of having been measured.

We measure disagreement against physical error directly. A \emph{probe} maps
latents to simulator state, so a prediction can be scored in pixels of block
displacement rather than in feature distance, and we report the probe's own
error per condition as the floor on everything measured through it.

We also separate two modes, because the signal is applied in one and the
evidence for it is gathered in the other. In \emph{open loop} the ensemble sees
one frame and an action sequence and feeds its own predictions back; this is
the mode a planner runs in when it scores an imagined rollout or cuts it short,
and the mode the penalties and gates above are applied to
\cite{mopo,lompo,safetyfilter}. In \emph{monitoring} it predicts one step from
the last three real frames, the only mode in which a change in the world
reaches its input at all, and the regime short-horizon evaluations of
disagreement sit closest to \cite{intervention}. Within those modes we apply
shifts of controlled severity (changed friction, a slippery floor, an
unobserved shove, a lighting change), extend the open-loop horizon to 50 steps,
and change what the model sees, to separate what the frozen features carry from
what the predictor uses.

Our contributions:
\begin{enumerate}\itemsep2pt
 \item \textbf{Physical disagreement measure.} A probe from latents to
 simulator state, so disagreement is compared against pixels of block
 displacement rather than feature distance, with the probe's own error
 reported per condition as the floor on everything measured through it,
 and with open loop separated from monitoring because the signal is
 applied in one and its evidence gathered in the other.
 \item \textbf{Interpretation of the measure.} Disagreement estimates difficulty; it does not detect novelty. It ranks physical error one step ahead inside
 the training distribution, decays to near chance over an imagined
 rollout, and responds to no physical change we apply, falling under the
 two largest, while rising most for a change that alters no physics at
 all. What it follows instead is how much the block is moving. The
 one-step observed residual, measured identically, does react.
 \item \textbf{Causal analysis.} The predictor reads the content of past frames while ignoring the block's response to its own actions, established by intervening on
 its input and reproduced on the released checkpoint. No statistic over
 its predictions can recover what those predictions ignore, so we do not
 expect a more diverse ensemble to be the remedy.
\end{enumerate}

\noindent The remainder of the paper is structured as follows: In Section 2, we
cite some of the most relevant and adjacent works. Section 3 lays out the setup
used to conduct the experiments. Section 4 presents the results, with some more
supporting experiments in Section 5. In Section 6, we briefly discuss the
findings and their implications. We conclude by stating the limitations of this
study in Section 7 and the possible next steps in Section 8.

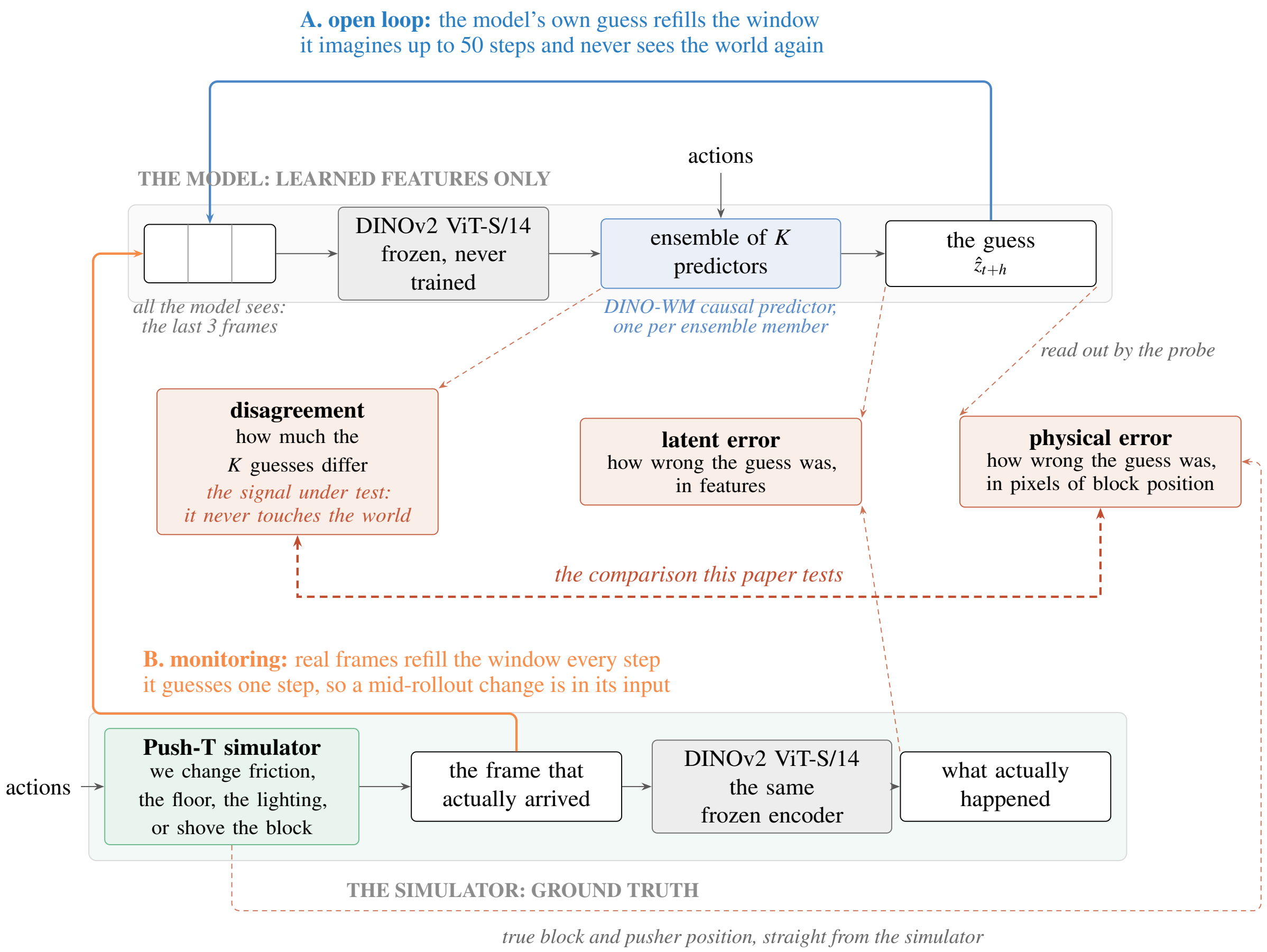
\begin{figure}[t]
\centering
\resizebox{\textwidth}{!}{% Overview schematic, functional labelling, named components.
\hyphenpenalty=10000\exhyphenpenalty=10000
\begin{tikzpicture}[
  x=1cm, y=1cm, font=\small,
  box/.style={draw, rounded corners=2pt, minimum height=9mm, align=center,
              fill=white, text width=2.6cm, inner sep=4pt, font=\footnotesize},
  frozen/.style={box, fill=black!7, draw=black!55},
  pred/.style={box, draw=NavyBlue!70, fill=NavyBlue!7, text width=3.0cm},
  world/.style={box, draw=ForestGreen!60, fill=ForestGreen!8, text width=3.2cm},
  meas/.style={draw=BrickRed!75, fill=BrickRed!6, rounded corners=2pt,
               align=center, text width=3.5cm, inner sep=5pt, font=\footnotesize},
  sub/.style={font=\scriptsize\itshape, text=black!55, align=center},
  flow/.style={-{Stealth[length=4pt]}, draw=black!65},
  modeA/.style={-{Stealth[length=4pt]}, draw=NavyBlue!75, line width=0.9pt},
  modeB/.style={-{Stealth[length=4pt]}, draw=Orange!90, line width=0.9pt},
  into/.style={-{Stealth[length=3.5pt]}, draw=BrickRed!65,
               dash pattern=on 2pt off 1.6pt},
  ask/.style={{Stealth[length=4.5pt]}-{Stealth[length=4.5pt]}, draw=BrickRed!85,
              line width=0.9pt, dash pattern=on 3pt off 2pt},
  lane/.style={font=\scriptsize\bfseries, text=black!45},
]

% ---------- the slot the two modes fill differently ----------------------
\node[draw, rounded corners=2pt, minimum width=1.8cm, minimum height=8mm,
      inner sep=0pt, fill=white] (hist) at (2.2,0) {};
\draw[black!40] (1.9,-0.4) -- (1.9,0.4);
\draw[black!40] (2.5,-0.4) -- (2.5,0.4);
\node[sub] at (2.2,-0.88) {all the model sees:\\[-2pt] the last 3 frames};

% ---------- model lane ---------------------------------------------------
\node[frozen] (enc)  at (5.4,0)  {DINOv2 ViT-S/14\\frozen, never trained};
\node[pred]   (pred) at (9.2,0)  {ensemble of $K$\\predictors};
\node[box]    (zhat) at (12.9,0) {the guess\\[-2pt]{\scriptsize $\hat z_{t+h}$}};
\node[sub, text=NavyBlue!70] at (9.2,-0.88) {DINO-WM causal predictor,\\[-2pt] one per ensemble member};

\draw[flow] (hist) -- (enc);
\draw[flow] (enc) -- (pred);
\draw[flow] (pred) -- (zhat);
\node[font=\footnotesize] (act) at (9.2,1.35) {actions};
\draw[flow] (act) -- (pred);

\draw[modeA, rounded corners=3pt] (zhat.north) -- ++(0,1.9) -| (hist.north);
\node[font=\footnotesize, text=NavyBlue!85, anchor=south west, align=left] at (3.3,2.55)
  {\textbf{A. open loop:} the model's own guess refills the window\\[-1pt]
   it imagines up to 50 steps and never sees the world again};

% ---------- world lane ---------------------------------------------------
\node[world] (sim)   at (2.5,-7.3) {\textbf{Push-T simulator}\\[-2pt]{\scriptsize we change friction, the floor, the lighting, or shove the block}};
\node[box]   (otrue) at (6.4,-7.3) {the frame that\\actually arrived};
\node[frozen, text width=3.0cm](enc2)  at (9.9,-7.3) {DINOv2 ViT-S/14\\the same frozen encoder};
\node[box]   (ztrue) at (13.1,-7.3) {what actually\\happened};

\node[font=\footnotesize] (act2) at (-0.15,-7.3) {actions};
\draw[flow] (act2) -- (sim.west);

\draw[flow] (sim) -- (otrue);
\draw[flow] (otrue) -- (enc2);
\draw[flow] (enc2) -- (ztrue);

\draw[modeB, rounded corners=3pt]
  (otrue.north) -- (6.4,-6.3) -- (0.6,-6.3) -- (0.6,0) -- (hist.west);
\node[font=\footnotesize, text=Orange!90, anchor=north west, align=left] at (1.15,-5.3)
  {\textbf{B. monitoring:} real frames refill the window every step\\[-1pt]
   it guesses one step, so a mid-rollout change is in its input};

% ---------- measurement band ---------------------------------------------
\node[meas] (dis) at (3.4,-2.85)
  {\textbf{disagreement}\\[-1pt]{\scriptsize how much the $K$ guesses differ}\\[1pt]
   {\scriptsize\itshape\color{BrickRed!80} the signal under test:\\[-2pt] it never touches the world}};
\draw[into] (pred.south west) -- (dis.north east);

\node[meas] (lat) at (9.2,-2.85)
  {\textbf{latent error}\\[-1pt]{\scriptsize how wrong the guess was,\\[-2pt] in features}};
\draw[into] (zhat.south west) -- (lat.north east);
\draw[into] (ztrue.north west) -- (lat.south east);

\node[meas] (phys) at (14.4,-2.85)
  {\textbf{physical error}\\[-1pt]{\scriptsize how wrong the guess was,\\[-2pt] in pixels of block position}};
\draw[into] (zhat.south east) -- node[pos=0.5, right=1pt, font=\scriptsize\itshape,
  text=black!60] {read out by the probe} (phys.north west);

% the simulator's own state is the ground truth; no second box for it
\draw[into, rounded corners=3pt] (sim.south) |- (9.5,-9.0) -| (16.6,-2.85)
  -- (phys.east);
\node[font=\scriptsize\itshape, text=black!60, anchor=north] at (9.5,-9.12)
  {true block and pusher position, straight from the simulator};

% ---------- the question the paper asks ----------------------------------
\draw[ask] (dis.south) -- (3.4,-4.7) -- node[above, font=\footnotesize\itshape,
  text=BrickRed!85] {the comparison this paper tests} (14.4,-4.7) -- (phys.south);

% ---------- lanes --------------------------------------------------------
\begin{scope}[on background layer]
  \node[fit=(hist)(zhat), draw=black!15, rounded corners=3pt,
        inner sep=6pt, fill=black!2] (modelbg) {};
  \node[fit=(sim)(ztrue), draw=black!15, rounded corners=3pt,
        inner sep=6pt, fill=ForestGreen!4] (worldbg) {};
\end{scope}
\node[lane, anchor=south west] at ([shift={(0mm,1.2mm)}]modelbg.north west)
  {THE MODEL: LEARNED FEATURES ONLY};
\node[lane, anchor=north west] at ([shift={(34mm,-1.6mm)}]worldbg.south west)
  {THE SIMULATOR: GROUND TRUTH};

\end{tikzpicture}}
\caption{The measurement setup. The two modes differ only in what refills the
model's three-frame window: its own output in \textbf{A}, real frames in
\textbf{B}. Only in \textbf{B} is a mid-rollout change present in the model's
input, so every intervention is run there. Disagreement and latent
error are computed entirely inside the learned feature space; only the probe
crosses into physical units, and only the simulator supplies ground truth.}
\label{fig:overview}
\end{figure}

% ============================================================
\section{Related work}\label{sec:related}
% ============================================================

\paragraph{World models on frozen features.}
DINO-WM \cite{dinowm} learns a causal predictor over frozen DINOv2 patch tokens
from offline trajectories, with no rewards or demonstrations, and plans to new
goals at test time. DINO-world \cite{dinoworld} scales the same approach and names
the deterministic predictor as a known limitation. V-JEPA~2-AC \cite{vjepa2}
attaches an action-conditioned predictor to an internet-scale video model. All
three predict point estimates.

\paragraph{Ensemble disagreement as uncertainty.}
Disagreement among dynamics models is long established as an epistemic signal, in
low-dimensional state \cite{morel,mopo}, in learned image latents
\cite{lompo,plan2explore}, and in latent safety filters \cite{safetyfilter}.
Table~\ref{tab:prior} summarises what is ensembled, in what space, and to what
end.

\paragraph{Tests of the assumption in latent space.}
The two results that bear directly on ours are stated in
\S\ref{sec:intro}: \cite{biaseddreams} for RSSMs, and \cite{intervention} for
frozen DINOv2 predictors in feature space, which is that paper's closest
substrate to ours though not its only one; it also audits TD-MPC2, DreamerV3
and LeWorldModel checkpoints on DeepMind Control. A third line of work keeps the
latent space but changes the statistic: Cao and Bansal \cite{worldmodelslie}
note that disagreement can stay small while predictions deviate from
observations, and substitute conformal inference on \emph{observed} latent error, a measured quantity rather than a self-assessment, which is the
direction our last result also points.

\begin{table}[t]
\centering
\small
\caption{Prior uses of dynamics-model disagreement. The last three rows are
checked against their papers; the first five are the standard characterisations
of methods we do not re-derive here.}
\label{tab:prior}
\begin{tabular}{@{}llll@{}}
\toprule
Work & Ensembled over & Space & Use \\
\midrule
MOReL \cite{morel} & dynamics models & low-dim state & unknown-state detector \\
MOPO \cite{mopo} & probabilistic models & low-dim state & reward penalty \\
Plan2Explore \cite{plan2explore} & 1-step latent predictors & learned latent & exploration bonus \\
LOMPO \cite{lompo} & latent dynamics models & learned latent & offline-RL penalty \\
Safety filters \cite{safetyfilter} & latent world model & latent / features & trust gating \\
Berger et al.\ \cite{biaseddreams} & RSSM predictors & recurrent latent & vs compounding error \\
Vakalis \cite{intervention} & 5 predictors & frozen DINOv2 & rank feature error \\
\textbf{This work} & 3 predictors & frozen DINOv2 & vs \emph{physical} error \\
\bottomrule
\end{tabular}
\end{table}

% ============================================================
\section{Setup}\label{sec:setup}
% ============================================================

\subsection{Substrate}
We use DINO-WM \cite{dinowm} on Push-T. The encoder is a frozen DINOv2 ViT-S/14
producing $256 \times 384$ patch tokens per frame; the predictor is the released
causal architecture over a three-frame history. Each frame in that history
carries its patch tokens, the pusher's position and velocity as four explicit
numbers, and the action, five environment steps concatenated. The block's
position and motion appear nowhere in that list. They are available only in the
pixels, which matters for \S\ref{sec:mechanism}: the predictor is told how fast
its own arm is moving and has to read everything about the block from the
image. Features for 9{,}000 episodes (8{,}000 for ensemble
training, 1{,}000 held out) were cached once at every fifth frame, matching the
model's frameskip. The held-out split is our own, drawn with a fixed seed from
\texttt{train}, because the repository's \texttt{val} split contains only 21
episodes, and because \texttt{val} is also the source of the planning-eval
goals, so drawing our split elsewhere keeps the probe uncontaminated by them.

We reproduce \SI{64.7}{\percent} planning success ($33/51$ episodes,
\SI{95}{\percent} CI \numrange{52}{78}) against \SI{66.0 \pm 4.7}{\percent}
reported, on a partial protocol run.

\subsection{Ensemble}
Three predictors were trained on a 4{,}000-episode bootstrapped subset with
identical architecture, loss, optimiser and global batch to the original implementation, each
stopped by the same compute guard (capped at 25 epochs).
Held-out teacher-forced visual L2 was $0.0758$, $0.0726$ and $0.0704$ against
the released checkpoint's $0.0319$ (Fig.~\ref{fig:a1} in the appendix), on
roughly \SI{5}{\percent} of the original's optimiser steps and \SI{21}{\percent} of
its episodes. Every result that turns on this gap is checked against the
released checkpoint directly.

Figure~\ref{fig:a3} shows that members' latent error follows the same curve
shape as the released model, but at no fixed offset: the ratio runs between
$1.6\times$ and $2.8\times$ under dataset replay and in live simulation, and
open loop it decays from $1.9\times$ at step 1 to ${\approx}1.0\times$ by step
50 as the released model saturates too. Their drift behaviour is therefore that
of the released model with a horizon-dependent gap rather than a constant one, and in any case this does not
establish that their epistemic properties scale the same way.

\subsection{Physical probe}
A small regressor maps latents to pusher position, block position and block
angle, trained on a disjoint split against simulator state. It is the same kind
of readout head as the original's own pose head, whose Push-T weights are not
released. It is an instrument only: it is
trained on privileged simulator state, and reading a predicted block position
is uninformative about error until the true position arrives. \emph{Physical error} throughout is the distance between the probe's reading
of a predicted latent and the simulator's own state, in pixels of the
\SI{512}{px} simulator frame. For scale in that frame, the T-shaped block spans
\SI{120}{px} across its crossbar with limbs \SI{30}{px} thick, and the circular
pusher is \SI{30}{px} across, so a \SI{10}{px} error is about a third of a limb
width. Two variants appear in the controls of \S\ref{sec:trust}: the
probe's error on the latent of the frame that actually arrived, which is the
floor on everything measured through it, and the distance between its reading
of the prediction and its reading of that arrived frame, which cancels any bias
it shares between the two. Its accuracy, and the margin it leaves over the
effects we measure, are reported with the other controls in \S\ref{sec:works}.

\subsection{Rollouts and conditions}
All conditions are paired on identical initial states and action sequences.
Dataset replay reaches only ${\approx}24$ model steps (held-out episodes are
short), so 50-step rollouts run in the live simulator with a contact-seeking
scripted pusher that moves the block in \SI{100}{\percent} of rollouts.
Live renders are round-tripped through the dataset's video codec before
encoding, which reduces a measured feature gap from \numrange{0.029}{0.031} to
\numrange{0.007}{0.011}; we report that residual beside sim latent errors.
$n = 200$ paired rollouts per condition.

\paragraph{Epistemic versus aleatoric uncertainty.}
Ensemble disagreement is an \emph{epistemic} signal: large where the model
lacks the knowledge to predict, shrinking as that knowledge is supplied. It is
not meant to capture \emph{aleatoric} uncertainty, the irreducible randomness
of the process, which a deterministic point predictor cannot represent at all.
Our conditions are all epistemic by that definition, since the simulator stays
deterministic under each and a model trained on varying friction could learn
to identify it, so disagreement is being asked for exactly what it is for.
But the distinction is a property of the model and the information reaching it
rather than of the world \cite{hullermeier}: a quantity the model cannot
identify from its inputs behaves like irreducible noise to it, however
reducible it is in principle. That orders our conditions. Off-distribution
actions are identifiable, since actions are an input and only the region is
unfamiliar. Changed friction and a slippery floor are identifiable only from
motion across frames. An unobserved shove is not identifiable at all.
\S\ref{sec:mechanism} turns on which of these the predictor can actually read.

\paragraph{Reporting convention.} Physical quantities are medians over paired
windows. Response ratios are geometric means of paired ratios over a fixed window after the change, with a bootstrap 95\% interval; ratios are
multiplicative, so the mean is taken in logs. The paired distributions are heavy-tailed and the
choice matters: on the slippery floor the median block error is \SI{16}{px}
against a mean of \SI{25}{px}. Means are named as such where used. Intervals
are cluster bootstraps over rollouts (20{,}000 resamples), since steps within a
rollout are not independent.

\paragraph{Open loop and monitoring.}
In a paired \emph{open-loop} rollout the ensemble sees only frame 0 and the
action sequence, so a mid-rollout change alters reality but neither input, and
its predictions are identical by construction. Shift experiments therefore run
in \emph{monitoring} mode, one step ahead from the last three observed frames
exactly as the members were trained, which makes available to it the
evidence a response would require. Open loop is what a planner runs on when it scores an imagined rollout or cuts
it short; monitoring is the only setting in which a mid-run change reaches the
model's input at all. Figure~\ref{fig:overview} sets out both modes and shows
where each quantity in this paper is measured, and which of them cross from the
learned space into physical units. Every
result below states its mode.

\paragraph{Friction.} We find that the original implementation sets \texttt{friction} on pymunk
\emph{bodies} rather than \emph{shapes}, so all Push-T training data has
frictionless pusher--block contact ($\mu = 0$). The sweep therefore uses
absolute $\mu \in \{0, 0.25, 0.5, 1, 2\}$, with a visible change in behaviour at
$\mu = 2$ where the pusher grips and the block translates without rotating.

% ============================================================
\section{Results}
% ============================================================

Disagreement does not register a changed world. It falls under the largest
physical change we induce, rises most for a change that alters no physics at
all, and never leaves the false-alarm rate as a detector. Where the features carry the change, as they do for the slippery floor and the lighting, the limit lies in the predictors rather than the representation: each reads where the block is while ignoring how it responded.

This section is the result: what disagreement fails to register
(\S\ref{sec:changed}), why it fails (\S\ref{sec:mechanism}), and what it does
measure instead and over what horizon (\S\ref{sec:works}). Five supporting
experiments substantiate it, stated briefly in \S\ref{sec:trust}, with the
numerical detail in Appendix~\ref{sec:trustdetail} and the figures in
Appendix~\ref{sec:appendix}. The strongest rescores every condition against the
probe's own reading of the true frame, which removes measurement bias and makes
the slippery-floor effect larger.

\subsection{Disagreement does not register a changed world}
\label{sec:changed}

The expectation under test is a strong one. The predictor's whole job is the
map from a latent and an action to the next latent, so a change in that map is
the thing it ought to be most sensitive to, and spread among predictors that
disagree about the map ought to be where the change shows up.

\begin{figure}[t]
\centering
\includegraphics[width=\textwidth]{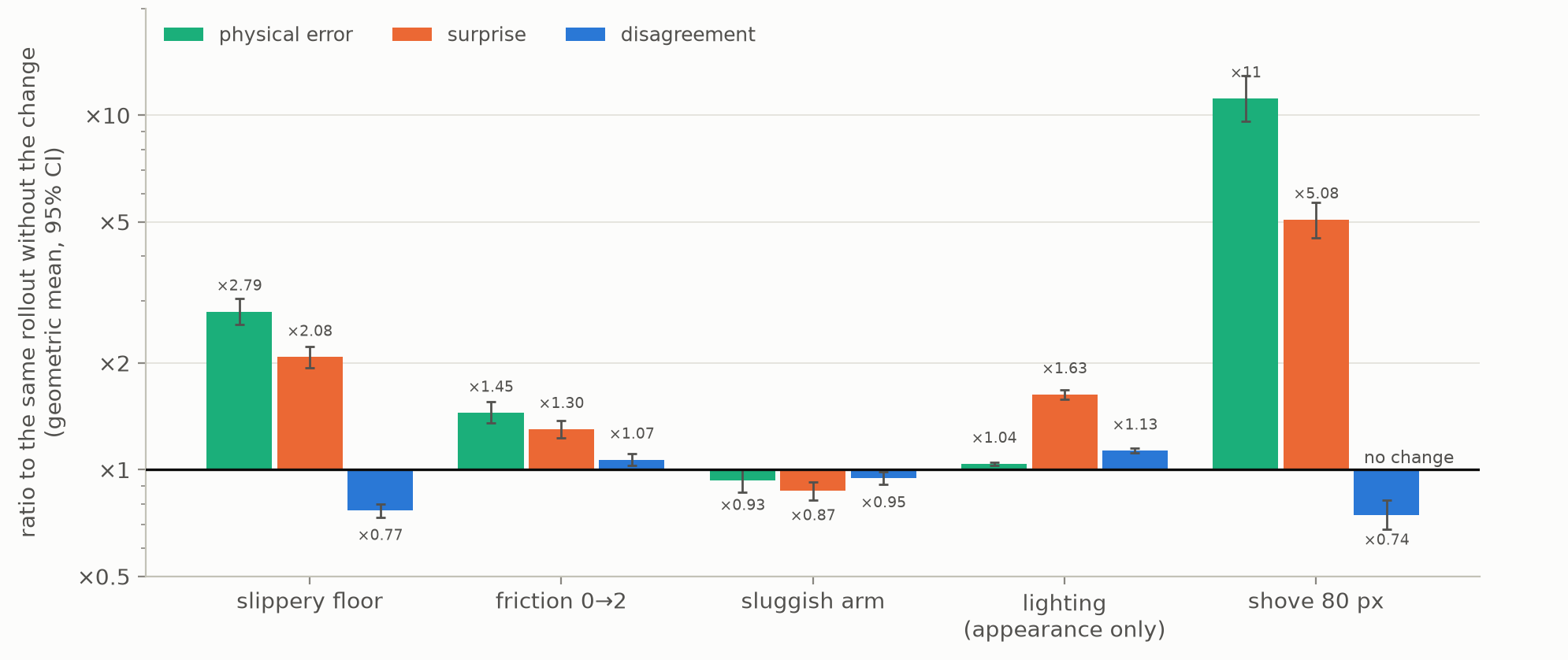}
\caption{Monitoring mode, 200 paired rollouts. Each bar is the ratio to the
same rollout without the change. Physical error and surprise rise with every
physical change; disagreement falls under the two largest and rises most for
the one change that alters nothing physical.}
\label{fig:b8}
\end{figure}

Each condition is a persistent change applied mid-rollout, scored one step
ahead from three observed frames over 200 held-out rollouts, against the ratio to the identical
rollout without it (Fig.~\ref{fig:b8}; the friction and shove severity sweeps are Fig.~\ref{fig:b3}). ``Surprise'' is the
one-step observed residual, the prediction against the frame that arrived.

\begin{itemize}\itemsep2pt
 \item \textbf{Slippery floor}, the block sliding on instead of stopping:
 physical error $\times 2.79$, surprise $\times 2.08$, and disagreement
 \emph{falls} to $\times 0.77$, flagging \SI{2}{\percent} of rollouts
 against its own \SI{5}{\percent} false-alarm rate.
 \item \textbf{Mid-rollout friction switch} $0 \to 2$: physical $\times 1.45$, surprise
 $\times 1.30$, disagreement $\times 1.07$.
 \item \textbf{\SI{80}{px} shove}, which the model had no way to anticipate:
 physical error $\times 11.2$ and surprise $\times 5.08$ at the step it enters the target but not yet the model's input, where disagreement is $\times 1.00$. Two steps on, with the displaced block
 squarely inside the input window, it is $\times 0.74$ and flags
 \SI{3}{\percent}.
 \item \textbf{Sluggish arm}: no quantity responds.
 \item \textbf{Lighting}, which alters no physics whatever: disagreement
 $\times 1.13$, its largest response in the set.
\end{itemize}

 Gated as an alarm, the signal fires less often
when the world has changed than when it has not: \SI{2}{\percent} of rollouts
on the slippery floor and \SI{3}{\percent} after the shove, against the
\SI{5}{\percent} it flags when nothing has changed. As a measure of
uncertainty, it tracks the scene's appearance and ignores its physics, so a
change in the dynamics registers as no uncertainty at all.

\paragraph{Correlation with block motion.}
Disagreement rises with how much the block is moving: Spearman
$\rho = 0.51$ against its angular velocity and $0.57$ against its
displacement, stable at every friction level
(Fig.~\ref{fig:b5}). But block speed is a property of the scene, whereas uncertainty is a property of the model, so a signal that follows it is not measuring
uncertainty.

That correlation is also a confound for everything above, because each
condition changes how much the block moves as well as how it moves. The test is
to compare only rollouts in which the block moved by the same amount, so motion
cannot account for the difference. Disagreement falls across the static friction sweep as
friction rises, which is the kind of severity response the sweep was built to
look for. But higher friction also leaves the block moving less, and at equal motion the fall
vanishes, whether by partial correlation ($\rho = +0.007$ against a raw
$-0.095$) or by comparing only steps of comparable displacement, where most of the fall
disappears (Fig.~\ref{fig:b5}). The motion
correlation accounts for that sweep and we do not count it as evidence.

The slippery floor does the opposite: there the block moves \emph{more} (angular velocity $\times 1.85$, displacement $\times 1.94$), so motion
alone predicts a rise, and disagreement falls to $\times 0.77$. Matching on
motion strengthens the effect, taking the partial rank
correlation from $-0.19$ to $-0.41$, with a fall in every quintile of equal
displacement (Appendix~\ref{sec:trustdetail}).

So the signal follows how much is happening, and departs from that pattern only
to fall when the physics breaks and rise when the pixels change. So it estimates difficulty and misses novelty. \S\ref{sec:discussion} returns
to why a quantity tracking input density should behave this way.

\subsection{The predictor does not read the block's motion}
\label{sec:mechanism}

Two explanations remain. Either the frozen features do not carry these changes,
or they do and the predictor ignores them. Both tests below point at the predictor.

\paragraph{The features carry the change.}
A small classifier distinguishes a post-switch three-frame window from the same
rollout unswitched, the same windows the world model is given, at AUROC
$0.77$ $[0.71, 0.83]$ for the slippery floor and $1.00$ for lighting, with all
pre-switch controls at $0.50$. Plotting that decodability against the
disagreement response (Fig.~\ref{fig:b9} in the appendix) shows no
relationship: the perfectly
decodable change produces the largest rise, and the clearly decodable physical
one produces a fall. Availability in the features does not predict a response.
Friction and a sluggish arm are weak in the features themselves, at $0.60$ and $0.62$ against $0.88$ from
raw pixels, so for those the representation is a genuine constraint. For the slippery floor
it is not, and neither is the block itself: the probe recovers its position
from these frozen latents to \SI{2.7}{px} on held-out data
(Appendix~\ref{sec:trustdetail}), about a tenth of a limb width and far finer
than the \SIrange{22}{46}{px} of travel the motion-free intervention below
removes. Whatever the predictor fails to use, the encoder resolves it.

\paragraph{The predictor discards the motion.}
We then change what the model sees. The current frame, action and
proprioception stay real; the two history frames are either real,
\textbf{swapped} for another rollout's at the same timestep, or \textbf{static}, the current frame and its proprioception
repeated, which removes every motion cue in the history while leaving the scene and the
real actions intact. Across those three frames
the block travels a median \SI{22}{px} and turns \SI{7}{\degree} on the normal
floor, and \SI{46}{px} and \SI{17}{\degree} on the slippery floor once it is
sliding, where \SI{94}{\percent} of steps move more than \SI{3}{px}. We run
this on the normal floor and on the slippery floor, where one frame
underdetermines the next (40 rollouts, the released checkpoint and all three
members).

Removing the motion costs almost nothing, precisely where the motion is the information it lacks. On the slippery floor the static-against-real latent error ratio is
$0.975$ $[0.968, 0.983]$ for the released checkpoint, marginally \emph{better}
without the motion, and $1.017$--$1.028$ for the members. These are the
rollouts where the model is already wrong by a median \SI{16}{px} against
\SI{6}{px} on the normal floor, because the block is sliding and it does not
predict that: the motion it discards is the information it is missing.
Appendix~\ref{sec:trustdetail} gives the same null in physical units.

These are small effects rather than exact nulls, and the swap condition supplies the scale they sit against. Deleting the motion
changes one-step error by at most a few percent; substituting the scene changes
it by $6$--$65$\%. Whatever use the predictor makes of its history, it is an
order of magnitude more sensitive to what the frames contain than to how things
moved between them.

\begin{figure}[t]
\centering
\includegraphics[width=\textwidth]{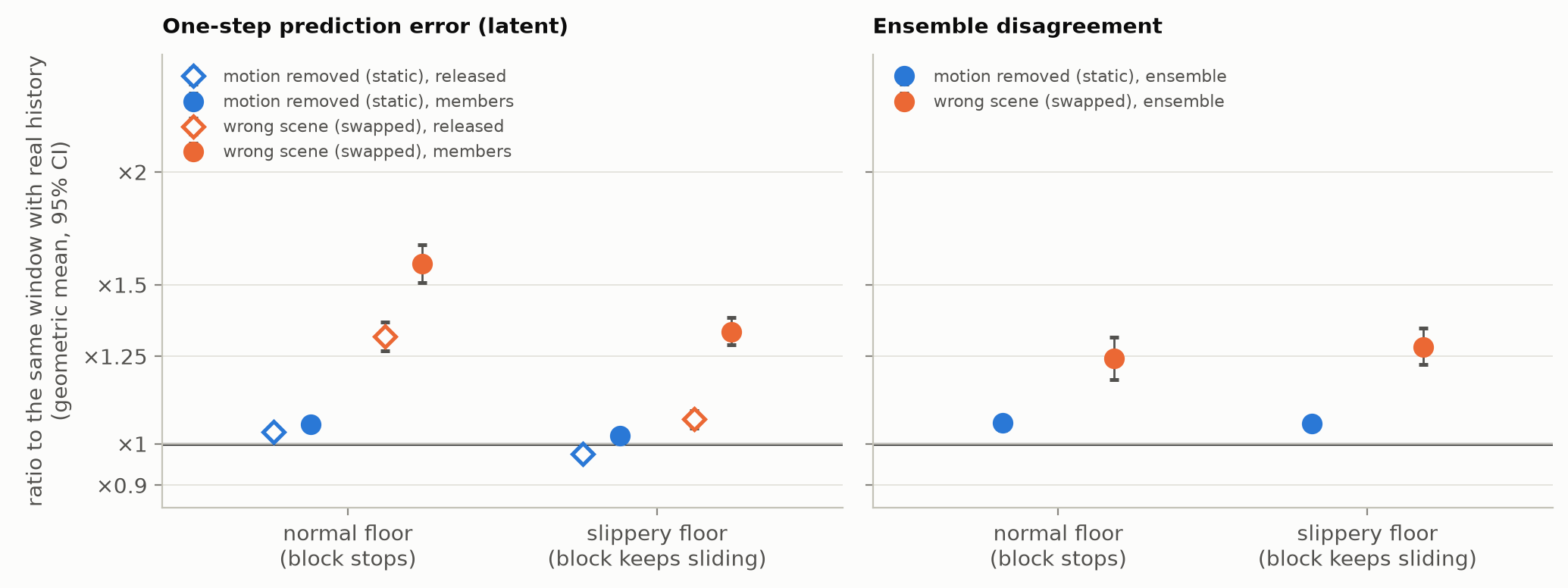}
\caption{Changing the frames the model sees. Removing the motion from the
history (blue)
leaves one-step error and disagreement essentially unchanged, including on the
slippery floor where the block is sliding and the model is missing it.
Substituting a different scene (orange) moves both. }
\label{fig:b11}
\end{figure}

The frames are nonetheless read, which gives the null above its force. Swapping in a
different scene shifts predictions by $9$--$37$\% of each model's error scale on
the slippery floor and $36$--$69$\% on the normal floor, raises latent error
$\times 1.06$--$1.65$ and raises disagreement $\times 1.24$--$1.28$
(Fig.~\ref{fig:b11}). The predictor reads the content of past frames and ignores the motion between them.

Nothing about the architecture forced this. The predictor is given a three-frame history, the original's own
training context, and a second frame adds nothing to the first except what
changed between them, so the design itself commits to motion carrying
information. It is also handed the pusher's velocity outright, so the only
motion it has to work out for itself is the block's. \S\ref{sec:works} shows that commitment has lapsed: three
real frames rank physical error no better than one at any horizon, and the
released checkpoint behaves the same way. Capacity is not the constraint either,
since attention across frame positions is exactly how motion would be
computed. A change in dynamics shows up in one place only, how
the block responded across frames, and the predictor leaves it unread. No statistic over its predictions can reflect what those
predictions ignore, and disagreement follows suit: an implausible history
raises it, a motion-free one barely moves it ($\times 1.05$ $[1.04, 1.07]$,
small but reliably above $1$). It answers unfamiliar inputs and overlooks unfamiliar dynamics, which is the lighting false alarm of
\S\ref{sec:changed} again. \S\ref{sec:discussion} takes up what follows: why
the predictor had no reason to learn this, and why a more diverse ensemble would not
supply it.

\subsection{Disagreement measures difficulty, and decays with horizon}
\label{sec:works}

\begin{figure}[t]
\centering
\includegraphics[width=\textwidth]{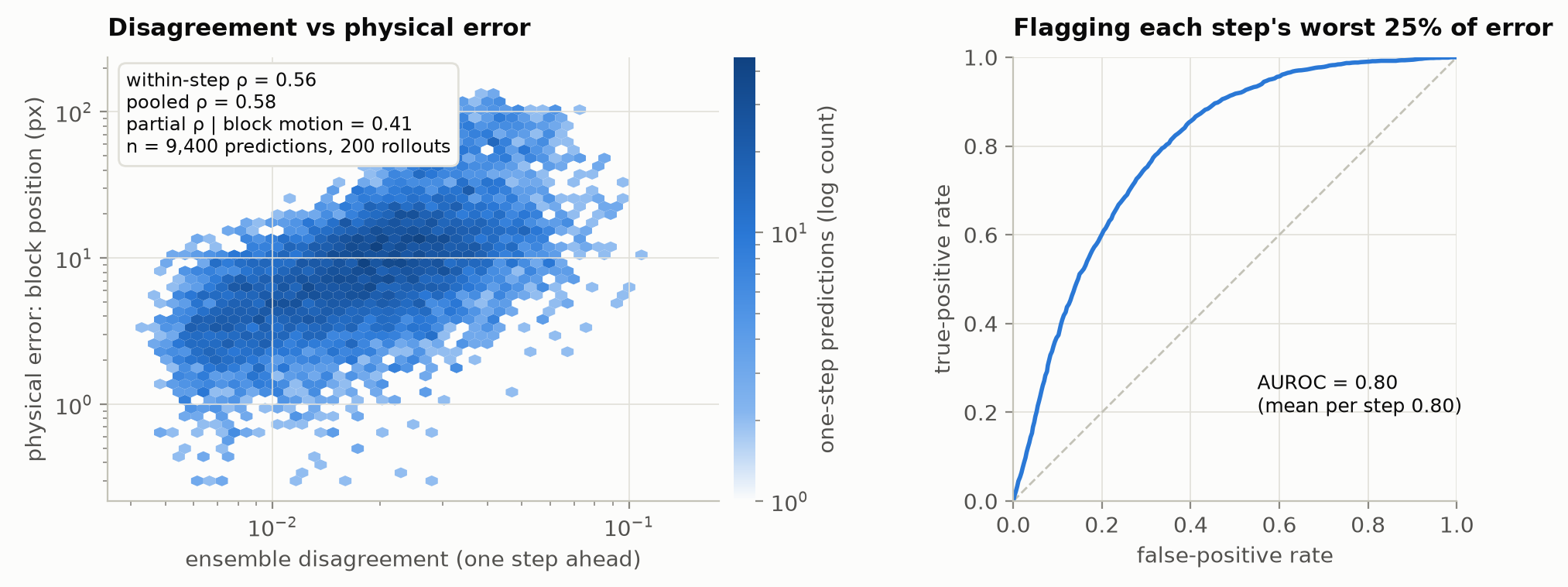}
\caption{Monitoring mode, in distribution. Disagreement ranks one-step physical
error at $\rho = 0.56$ within step ($0.41$ holding block motion fixed), and
flags each step's worst quartile of error at AUROC $0.80$.}
\label{fig:b7}
\end{figure}

The signal is not uninformative everywhere. Predicting one step ahead inside
the training distribution (200 rollouts of 47 post-history steps, 9{,}400 predictions), disagreement
ranks physical error at within-step $\rho = 0.56$ and flags each step's worst
quartile at AUROC $0.80$ (Fig.~\ref{fig:b7}). This is not an artefact of the
block moving: holding block displacement fixed gives a partial $\rho$ of
$0.41$, with $0.40$ on moving steps alone and $0.54$ on static ones. Within
what it was trained on, and one step out, the ensemble does know which
situations it handles badly.

\begin{figure}[t]
\centering
\includegraphics[width=0.84\textwidth]{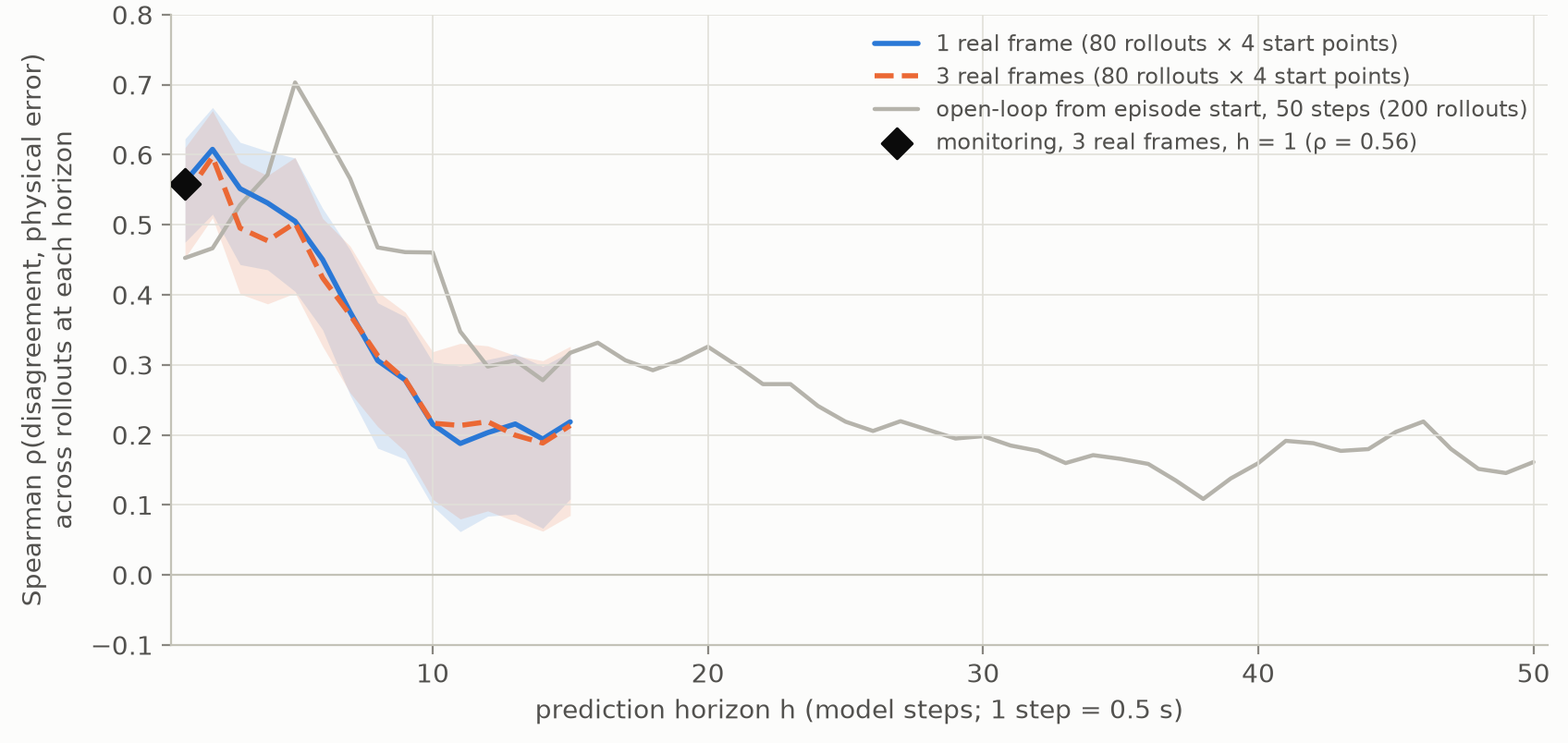}
\caption{The ranking decays with prediction horizon, and one real frame is
indistinguishable from three at every horizon. The grey reference curve runs
open-loop from the episode start with the original two-frame window, so its
first steps are not directly comparable.}
\label{fig:b10}
\end{figure}

That is the best case, and it decays. Over an imagined rollout $\rho$ falls from $0.56$ at $h = 1$ to $0.22$ by $h = 15$ (Fig.~\ref{fig:b10}), and the within-step AUROC from $0.77$ to $0.62$. Pooling across steps inflates this: because disagreement
and error both grow with horizon, a step counter alone ranks physical error as
well as disagreement does in every condition (Fig.~\ref{fig:b2b} in the
appendix), while within a single step stratified $\rho$ is
\numrange{0.19}{0.28}. Over 50 steps in distribution disagreement rises roughly linearly rather
than flattening, so the attractor behaviour reported for recurrent models
\cite{biaseddreams} does not reproduce here; the ranking degrades; the magnitude keeps growing.

Observed history does not rescue this. Giving the model three real frames
instead of one leaves $\rho$ unchanged at every horizon ($0.56$ against $0.54$
at $h = 1$; $0.20$ against $0.21$ at $h = 11$--$15$) and block error unchanged
as well ($6.7$ against $6.8$~px at $h = 1$). The released checkpoint behaves
the same way, with one-step latent error $0.0588$ on one frame against $0.0605$
on three and three frames better in \SI{49}{\percent} of paired cases, so this is a property of the released
model rather than of our training budget. Horizon alone separates the two modes.

% ============================================================
\section{Supporting experiments}\label{sec:trust}
% ============================================================

Five experiments substantiate the result. Two ask whether our own measurement
could have produced it, and both answer no; one of them makes the main effect
larger. Two check that the comparison was sound. The last marks which
conditions carry no information, so that they are not read as evidence either
way. Each is stated here in brief, with the numerical detail in
Appendix~\ref{sec:trustdetail}.

\paragraph{Probe accuracy is adequate in the conditions where it is used.} Its error is a property
of the condition rather than a constant, running from \SI{2.3}{px} under
dataset replay to \SI{16.6}{px} by step 50 open loop. In the monitoring
conditions where every shift experiment runs it reads \SI{4.8}{px} against a
median block error of \SI{7.9}{px}, so the one-step comparisons rest on pairing
rather than on headroom: each condition is scored against the same window, so
probe bias is the same on both sides and cancels.

\paragraph{Removing probe bias makes the effect larger.}
Rescoring every condition against the probe's own reading of the true frame
leaves the slippery floor \emph{stronger} ($\times 3.04$ bias-free against
$\times 2.79$ raw) and the shove stronger still ($\times 13.9$ against
$\times 11.2$ raw), and restricting the
ranking results to clean rows raises the partial $\rho$. Measurement error attenuates what we report; it does not create it. The friction sweep does not survive this, and we report
it as an explained negative (Fig.~\ref{fig:b3}).

\paragraph{The paired arms were identical until the change.} Every condition
matches its own unswitched pair to within $0.001$ on physical error, surprise
and disagreement before the switch, and every pre-switch decodability control
sits at $0.500$.

\paragraph{Three members and one statistic carry these conclusions.} Every
member pair and the full triple agree, and a MOReL-style max-pairwise statistic
\cite{morel} ranks rollouts almost identically to the spread measure used
throughout (rank agreement $0.993$). Absolute disagreement levels would still
shift with more members (\S\ref{sec:limits}).

\paragraph{Some conditions carry no information, and we do not read them.}
Where a change is applied that the ensemble never observes, its input is
identical with and without it and its predictions are identical by
construction. The two open-loop conditions that do carry information are
reported above and in the appendix: off-distribution actions raise physical
error $\times 2.6$ by step 10 while disagreement rises $\times 1.2$, and
compounding error in distribution grows disagreement without making it rank.

% ============================================================
\section{Discussion}\label{sec:discussion}
% ============================================================

\paragraph{Known unknowns only.}
In the usual terms, this ensemble detects known unknowns only. Ensembles are introduced as novelty detectors
\cite{morel,mopo,plan2explore}, and this one is, for novelty in
the input. What it misses is novelty in the dynamics, which is the kind a robot
meets and the kind a latent safety filter is deployed to catch: its purpose is
scenarios the world model has never encountered \cite{safetyfilter}, while its
signal is a statement about the familiarity of what it reads.

\paragraph{Covariate shift and concept shift.}
The standard taxonomy of distribution shift separates a change in the input
distribution $p(x)$ from a change in the conditional $p(y \mid x)$ with $p(x)$
left intact, the second of these being concept shift \cite{datasetshift}. A
change in dynamics is concept shift almost by definition: the same latent and
action now lead somewhere else, while the distribution of latents and actions
the model is asked about barely moves. Ensemble spread is a statement about
$p(x)$, so it is the wrong kind of instrument,
and every response we measure is consistent with this once sorted this way. The
lighting change is pure covariate shift and draws the largest response.
Off-distribution action sequences are covariate shift in one input channel and
draw a weak one. Changed friction, the slippery floor, the shove and the
sluggish arm are concept shift and draw nothing, or less than nothing. The
motion-magnitude correlation is no longer a separate finding either: fast and
rotating states are rarer in the training data, so a quantity tracking input
density ought to rise with motion, and the difficulty estimate of
\S\ref{sec:works} is that quantity seen from inside the training distribution.

\paragraph{Benchmarks for uncertainty under shift.}
The benchmarks make the same substitution. The two standard tests of predictive
uncertainty under shift corrupt the images \cite{canyoutrust} or move to a new
city, season or text domain \cite{shiftsbench}. Those are all changes to the
inputs. Ensembles do comparatively well on them, and uncertainty quality still
degrades as the shift grows. Our lighting change is the only condition here of
that kind, and it is the only one disagreement answers. Nothing in those
benchmarks changes the rule mapping inputs to outcomes, which changes when the floor turns slippery.

\paragraph{Not a new failure, a new setting for it.}
None of this is new in general. That a density estimate misses a change in the
conditional is well understood. What is new is where the failure is consequential. Models built on
pretrained frozen features are the promising direction for robot world models,
since they train from offline video with no rewards or demonstrations and plan
to goals they were never shown \cite{dinowm,dinoworld,vjepa2}, and what their
uncertainty estimates are worth has had little attention. Their predictors come
deterministic and their features come frozen, so the estimate we test is the
only one there is to add. The general result therefore applies where no alternative estimate exists,
and what we measure is not that the estimate is imperfect but that it responds
inversely to the error it should track, and why.

\paragraph{Two mechanisms, one of them fixable.}
Only one of them is reachable by a better ensemble. The first is shared support: the members differ only through
bootstrapping, over one frozen encoder, one data distribution and one
architecture, which produces spread where data is thin but gives them no
reason to extrapolate \emph{differently} once the world leaves that
distribution. Diversity (different encoders, randomised priors) is the natural test of it, and if that were the whole story a more varied ensemble
would help.
The second is the identifiability argument of \S\ref{sec:mechanism}, and it is
the decisive one because it closes that route off. What a change in dynamics offers the model is evidence in the block's response, which the predictor
discards, so the
members are not withholding uncertainty they hold; there is none to express.
Diversity of initialisation acts on how members extrapolate from what they read, and cannot manufacture a distinction their predictions never carried.

\paragraph{Redundancy in the training data.}
What is missing is a reason to learn the block's motion. On the normal floor
the block moves only while the pusher is in contact, so the current frame and
the action nearly determine the next one and velocity is redundant; the
training data also carries a single friction value throughout
(\S\ref{sec:setup}), so a predictor that fits it perfectly need not encode
friction at all, since one fixed law of motion suffices. The slippery floor is
where that redundancy ends, and the predictor still does not look. Being a
dynamics model does not make a predictor a dynamics-identifying one, and the
difference stays invisible until the dynamics change. Training across frictions
would very likely fix this instance without touching the general case: whatever
is held fixed in training is what the predictor has no reason to encode, and
the next deployed world will differ in something else. That is not a defect of
this particular training set but the situation the signal is advertised for:
train in one world, act in another, and be told when the world has moved.

\paragraph{Inferring the dynamics from history.}
The remedy has to change what is read rather than who reads it. Rapid motor adaptation estimates a latent
encoding of friction, payload and terrain from roughly half a second of state
and action history, and corrects for it on a deployed legged robot
\cite{rma}; context-based meta-RL maintains a posterior over a latent task
variable inferred from recent experience \cite{pearl}, which is an explicit
uncertainty about the dynamics rather than a by-product of parameter spread.
Both read what the predictor here discards. Whether either restores a
usable disagreement signal in a frozen-feature world model is the test
\S\ref{sec:mechanism} sets up, and the one we would run next. Short of that,
the result suggests a cheap diagnostic for any such model before its
uncertainty is trusted: establish that the quantity one expects it to be
uncertain about is identifiable from the inputs it reads at all.

\paragraph{Practical consequences.}
A planner has two places to read this signal, and both fail. Planning open loop from a single frame, it cannot see a change in dynamics at all: the change never enters its input, so every member predicts the trajectory it would have predicted in the unchanged world, and their spread is unchanged with it. Replanning on real observations, it can see the change and reads it backwards. Disagreement-based penalties and horizon
gating are applied to imagined rollouts \cite{mopo,lompo}, where the signal
degrades with the very horizon it is meant to police; a safety filter is asked
precisely to notice a world it was not trained on \cite{safetyfilter}, where it
does not respond at all. The one-step observed residual, which compares a prediction against what
actually arrived, does react to physical change, which is the practical case
for grounding trust in measured quantities. Conformal treatment of that residual is already proposed
\cite{worldmodelslie}, and model-value inconsistency \cite{selfinconsistency}
is a second signal that needs no additional members. Neither is a
self-assessment of the kind that fails here.

\paragraph{Scope of the claim.}
We do not claim the methods in Table~\ref{tab:prior} fail. Their substrates and
disagreement measures differ, and MOReL and MOPO operate on low-dimensional
state. The claim is about the assumption those methods share, evaluated in a
frozen-feature world model and measured in physical terms.

% ============================================================
\section{Limitations}\label{sec:limits}
% ============================================================

The ensemble itself is the main threat. It has three members, so the absolute
disagreement levels would shift with more, though the conclusions do not
(\S\ref{sec:trust}). More seriously, those members were trained to roughly
\SI{5}{\percent} of the released checkpoint's optimiser steps, which makes them
weaker than the deployed substrate, and epistemic disagreement shrinks with
data. This bears on the absolute levels we report rather than on the result.
Every response is a ratio against the same rollout without the change
(\S\ref{sec:setup}), so a uniform shift in level cancels; the members' spread
does respond to its inputs where there is something to respond to, rising
$\times 1.13$ for the appearance change and $\times 1.24$--$1.28$ for a
substituted history, so it is not saturated noise; and the mechanism results
reproduce on the released checkpoint itself (\S\ref{sec:mechanism},
\S\ref{sec:works}). What would move with better-trained members are the
false-alarm and flag rates, which depend on level directly, and a dataset-size
control at fixed optimiser steps would quantify them.

The scope is narrow in three further ways. We test Push-T and DINO-WM only, so
nothing here establishes that the result holds for frozen features in general,
and V-JEPA~2-AC is the natural second substrate. The account we give also turns
on members sharing one frozen encoder, so world models with learned encoders
\cite{lompo} may behave differently. And the simulator is deterministic while
our members are point predictors, so we never exercise stochastic transitions
or sensor noise and say nothing about how disagreement behaves when the process
itself is random, the case probabilistic ensembles \cite{mopo,lompo} are built
for; our conditions are epistemic by construction, and \S\ref{sec:setup} and
\S\ref{sec:mechanism} are where the finer question sits, which of them the
model could identify from its inputs at all.

The probe is the weakest instrument here: its error is condition-dependent and in the monitoring
conditions sits within a factor of two of the effects measured against it, so
pairing and the bias-free rescoring of \S\ref{sec:trust} are what make those
comparisons safe, and a decoder-free measurement would be better. The
decodability classifier may exploit trajectory-divergence cues rather than the
dynamics change itself, which a control holding the trajectory fixed would
resolve. The motion result is one-step and single-task: \S\ref{sec:mechanism}
changes a three-frame window predicting one step ahead in Push-T, and whether a
predictor given a longer context, or trained on data in which dynamics vary,
would read the block's motion is untested. Finally, we use a single seed set
and a partial baseline reproduction (51 of the benchmark's 96 episodes).

% ============================================================
\section{Next steps}
% ============================================================

Two further controls would strengthen this. A dataset-size control, varying training episodes
at fixed optimiser steps, would determine the absolute disagreement levels and
with them the flag rates (\S\ref{sec:limits}). A trajectory-matched control for
the decodability classifier would settle that question.

Two questions follow the result. The architectural test, a predictor given the object's motion
rather than only its own, or able to infer the dynamics from a short history as
in \cite{rma}, is what
\S\ref{sec:mechanism} says a response to changed dynamics requires,
with a diversity test (different encoders, randomised priors) as the control
that isolates the ensemble's own contribution. And a second substrate
(V-JEPA~2-AC), without which this remains a statement about DINO-WM rather than
about frozen-feature world models in general.

% ============================================================

% ============================================================
\appendix
\section{Supporting experiments in detail}\label{sec:trustdetail}
% ============================================================

The five experiments of \S\ref{sec:trust} in full.

\paragraph{Detail held back from \S\ref{sec:changed} and
\S\ref{sec:mechanism}.} Within quintiles of equal block displacement on the
slippery floor, the disagreement ratio is $0.86$, $0.68$, $0.55$, $0.60$ and
$0.64$, a fall in every one, while physical error in the upper three rises
$\times 1.8$--$2.6$. The $0.5$--$0.7$~px median displacement of the
\emph{predicted} block between the real-history and motion-free conditions is not the same quantity as the error differences below. Median block error under the motion-free history moves from $16.4$ to $16.1$~px for the
released checkpoint, marginally better without the motion and in the same
direction as the latent ratio, and is unchanged at $16.8$~px for the ensemble
mean. Paired and
bootstrapped over rollouts, removing the motion changes block error by
$-0.10$~px $[-0.24, +0.03]$ for the released checkpoint and $+0.03$~px
$[-0.15, +0.23]$ for the ensemble mean, both intervals containing zero, while
substituting the scene changes it by $+0.51$~px $[+0.20, +0.83]$ and
$+1.33$~px $[+0.76, +1.91]$ respectively.

\paragraph{Probe accuracy.}
The probe reads \SI{2.7}{px} (block), \SI{2.7}{px} (pusher) and
\SI{0.58}{\degree} on held-out data, and \SI{2.3}{px} on true future frames
under dataset replay. That floor is a property of the condition it is used in, and in places it is larger than the quantity measured against
it. In the live-sim monitoring conditions where every shift experiment runs,
its error on \emph{true} frames is \SI{4.8}{px} median and \SI{8.5}{px} mean at
$\mu = 0$, against a median block error of \SI{7.9}{px}, so the
effect is only $1.6$ times the probe's own error. Open loop it grows with horizon, to \SI{6.2}{px} by step 10
and \SI{16.6}{px} by step 50, against errors $12$--$15\times$ larger from step
10 on but only $1.9\times$ at step 1. 
Two conditions make the probe's own error worse rather than merely noisy. In
the in-distribution sim condition the simulator's pusher leaves the
\SI{512}{px} frame in \SI{47}{\percent} of rows, rising to \SI{66}{\percent} by
step 50, so pusher quantities there are undefined; block quantities are
unaffected and every physical result in this paper is block-based. In open loop
the block itself leaves the frame in \SI{7.3}{\percent} of rows.

\paragraph{Probe bias does not produce the effect.}\label{sec:instrcontrol}
Rescoring each condition against the probe's reading of the true frame
removes the bias entirely and changes the picture in one place only. The slippery floor is strengthened: the probe's own error is flat across that
switch ($\times 0.93$ $[0.86, 1.00]$) and the bias-free block error is
$\times 3.04$ $[2.77, 3.32]$ against $\times 2.79$ raw, rising to
$\times 4.13$ $[3.56, 4.80]$ on rows where the block is in frame and the floor
is under \SI{5}{px}. The shove is strengthened too ($\times 13.9$ $[11.9, 16.3]$ bias-free against
$\times 11.2$ raw at \SI{80}{px}, the floor itself \emph{falling} to
$\times 0.80$ $[0.69, 0.93]$), scored at the step the shove enters the target as
in \S\ref{sec:changed}. The
mid-rollout friction switch survives at reduced size ($\times 1.35$
$[1.26, 1.45]$ against $\times 1.45$ raw, the floor itself up $\times 1.16$
$[1.07, 1.24]$). Intervals here are cluster bootstraps over the 200 rollouts,
the same set \S\ref{sec:changed} reports. The friction
\emph{sweep} does not survive, and neither does its disagreement decline once
block motion is held fixed (\S\ref{sec:changed}), so we report it as an
explained negative (Fig.~\ref{fig:b3}). Restricting every ranking result to rows where the
block is in frame and the floor is under \SI{3}{px} \emph{raises} the
horizon-controlled partial $\rho$, from $0.27$ to $0.35$ open loop and $0.36$
to $0.41$ in monitoring, and scoring the within-step ranking against the
bias-free target raises AUROC from $0.683$ to $0.712$ at step 1 open loop. Measurement
error attenuates the reported relationships; it does not create them.
The lighting change is the reference case. The simulator is byte-identical
across it, so its physical response is zero by construction and the
$\times 1.04$ we measure is the probe reading a relit image. Its disagreement
response, $\times 1.13$, is not explainable that way.

\paragraph{The paired conditions did not differ beforehand.}
Before the intervention step every condition matches its own unswitched
pair to within $0.001$ on physical error, surprise and disagreement, and every
pre-switch decodability control sits at $0.500$ exactly. The arms are identical
until the change is applied.

\paragraph{Member pairs and the max-pairwise statistic agree.}
Across 30 rollouts every member pair and the full triple support the same
conclusions (slippery floor $\times 0.78$--$0.84$ against physical $\times 2.52$ on that
smaller set; lighting indistinguishable from friction; shove detection at or
below the false-alarm level). A MOReL-style max-pairwise statistic
\cite{morel} gives $\rho = 0.38$ against $0.36$ and ranks rollouts almost
identically to the spread measure used throughout (rank agreement
$0.993$). Absolute levels would still shift with more members
(\S\ref{sec:limits}).

\paragraph{Uninformative conditions.}
Where a change is applied that the ensemble never observes, its input is
identical with and without the change and its predictions are identical by
construction, so those conditions are uninformative; this is why every shift
experiment above runs one step ahead from observed frames. Two open-loop
conditions do carry information. Off-distribution action sequences are an input
the model receives: physical error rises $\times 2.6$ by step 10 while
disagreement rises $\times 1.2$ and is \emph{lower} than in-distribution by
step 50. And compounding error in distribution needs no external change at all: as a rollout extrapolates, predictions enter regions the members saw little
of, where disagreement is expected to grow and to rank. It grows; it ranks
weakly.

% ============================================================
\section{Supporting figures}\label{sec:appendix}
% ============================================================

\begin{figure}[t]
\centering
\includegraphics[width=0.78\textwidth]{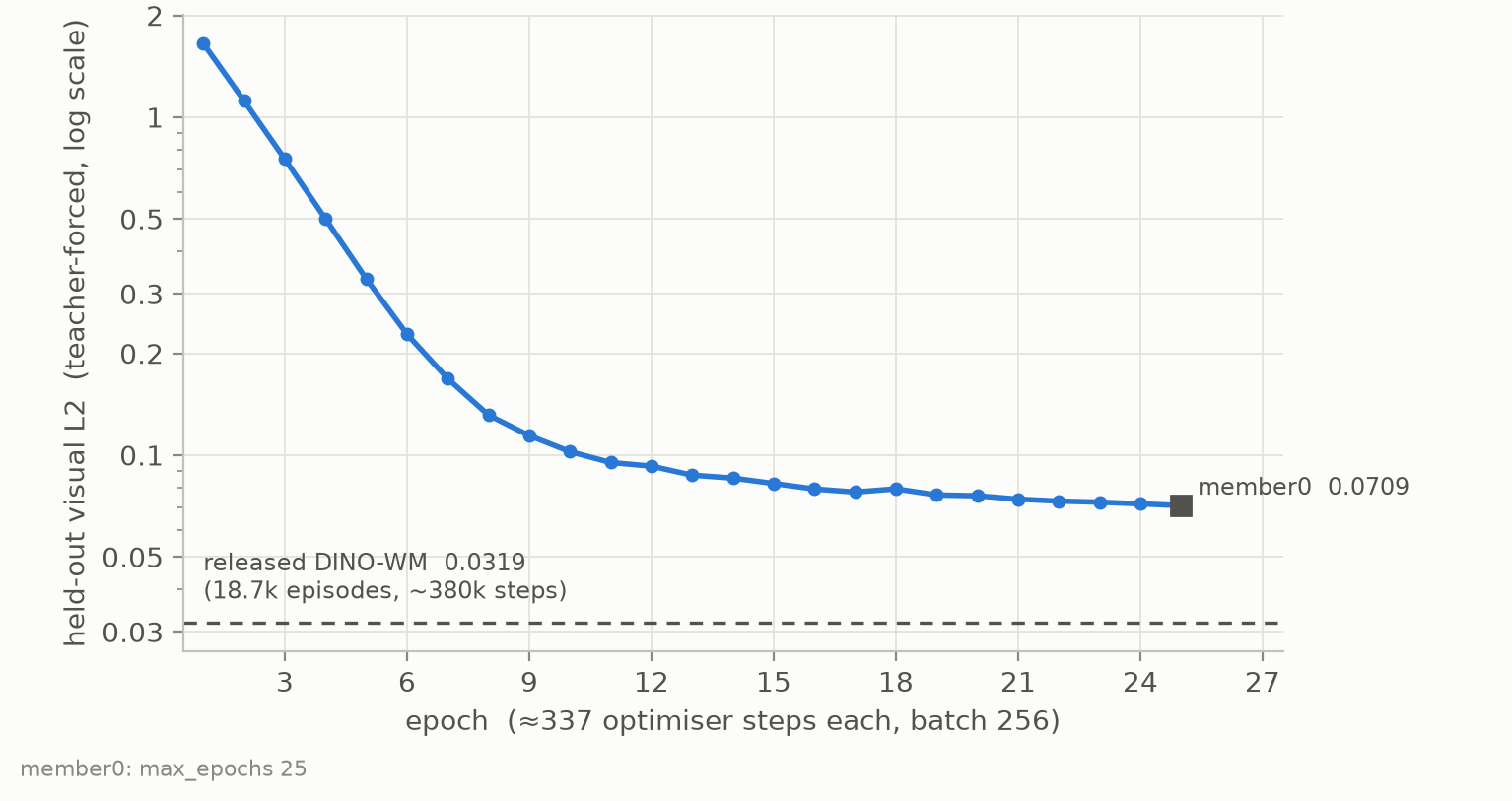}
\caption{Member 0 training. Held-out loss plateaus at $2.2\times$ the released
checkpoint after 25 epochs. The gap is the main threat to the results and is
addressed in \S\ref{sec:limits}.}
\label{fig:a1}
\end{figure}

\begin{figure}[t]
\centering
\includegraphics[width=0.9\textwidth]{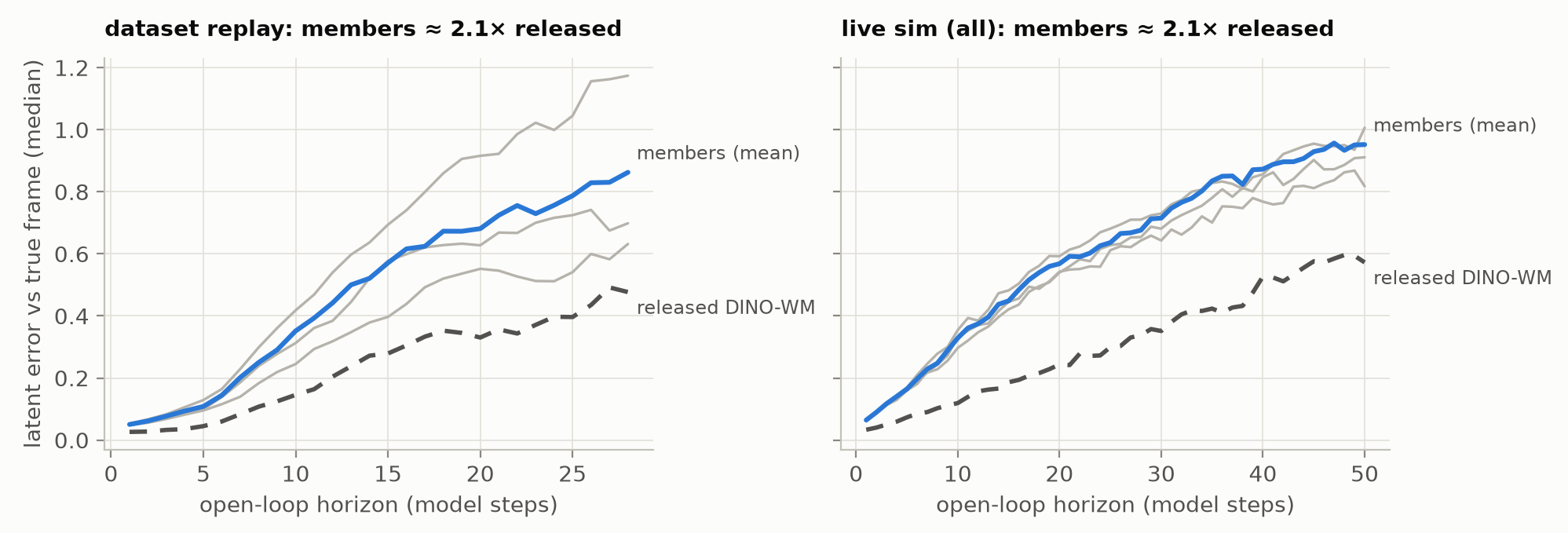}
\caption{Members drift with the same curve shape as the released model but at
no constant offset. The ratio of their latent error to its own runs $1.6$--$2.8\times$ under
replay and in live simulation, and falls from $1.9\times$ at step 1 to
${\approx}1.0\times$ by step 50 open loop as the released model saturates
too.}
\label{fig:a3}
\end{figure}

\begin{figure}[t]
\centering
\includegraphics[width=0.88\textwidth]{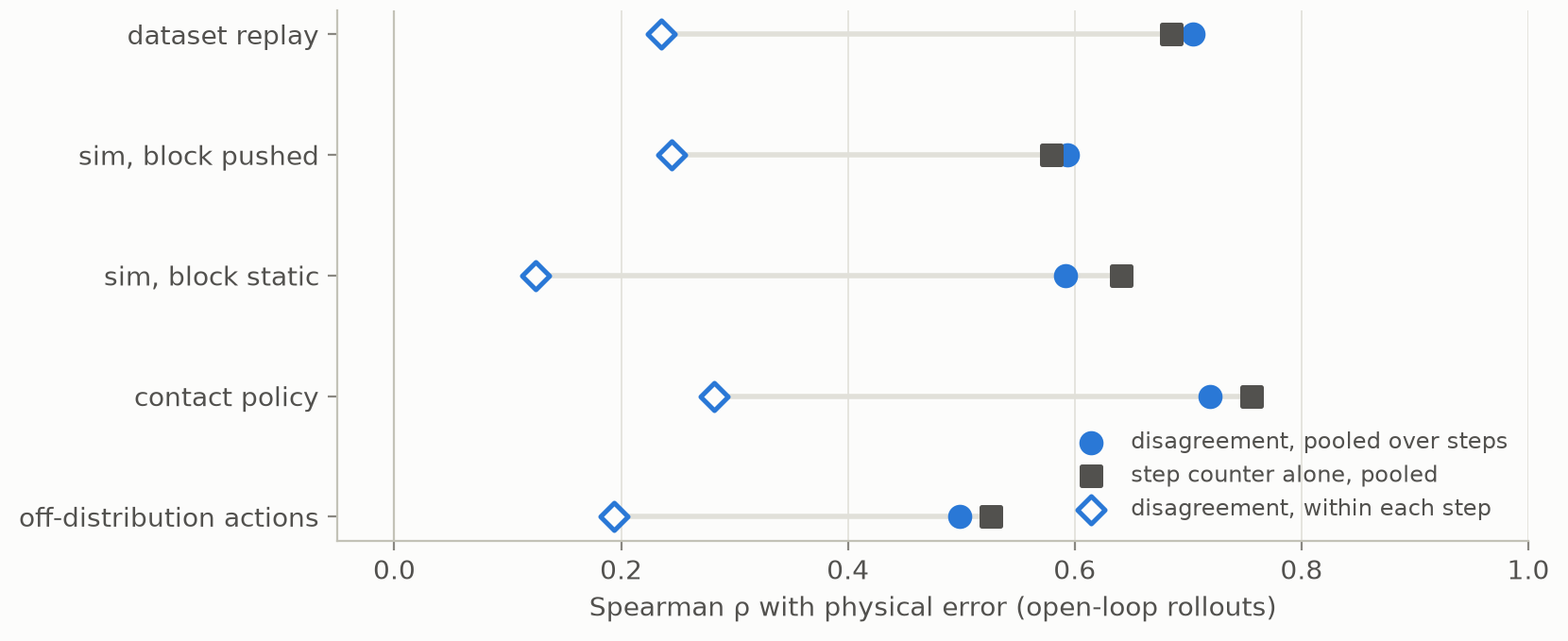}
\caption{Pooling over steps inflates the correlation. Elapsed step number
alone (grey) ranks physical error as well as disagreement does (blue circles);
within a single step, disagreement retains much less (open diamonds).}
\label{fig:b2b}
\end{figure}

\begin{figure}[t]
\centering
\includegraphics[width=0.78\textwidth]{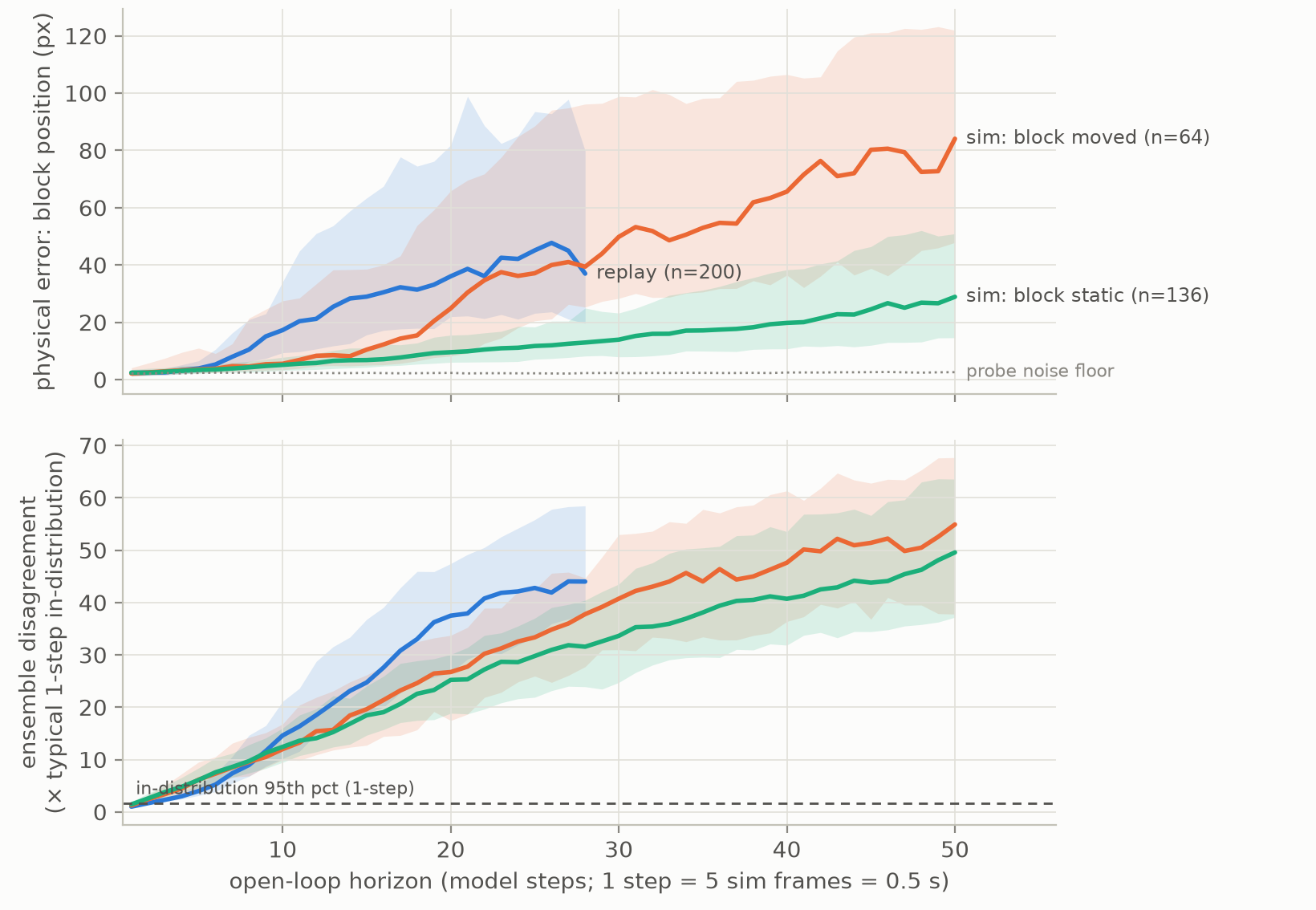}
\caption{Physical error separates sharply by whether the block moved;
disagreement barely does.}
\label{fig:b1}
\end{figure}

\begin{figure}[t]
\centering
\includegraphics[width=0.92\textwidth]{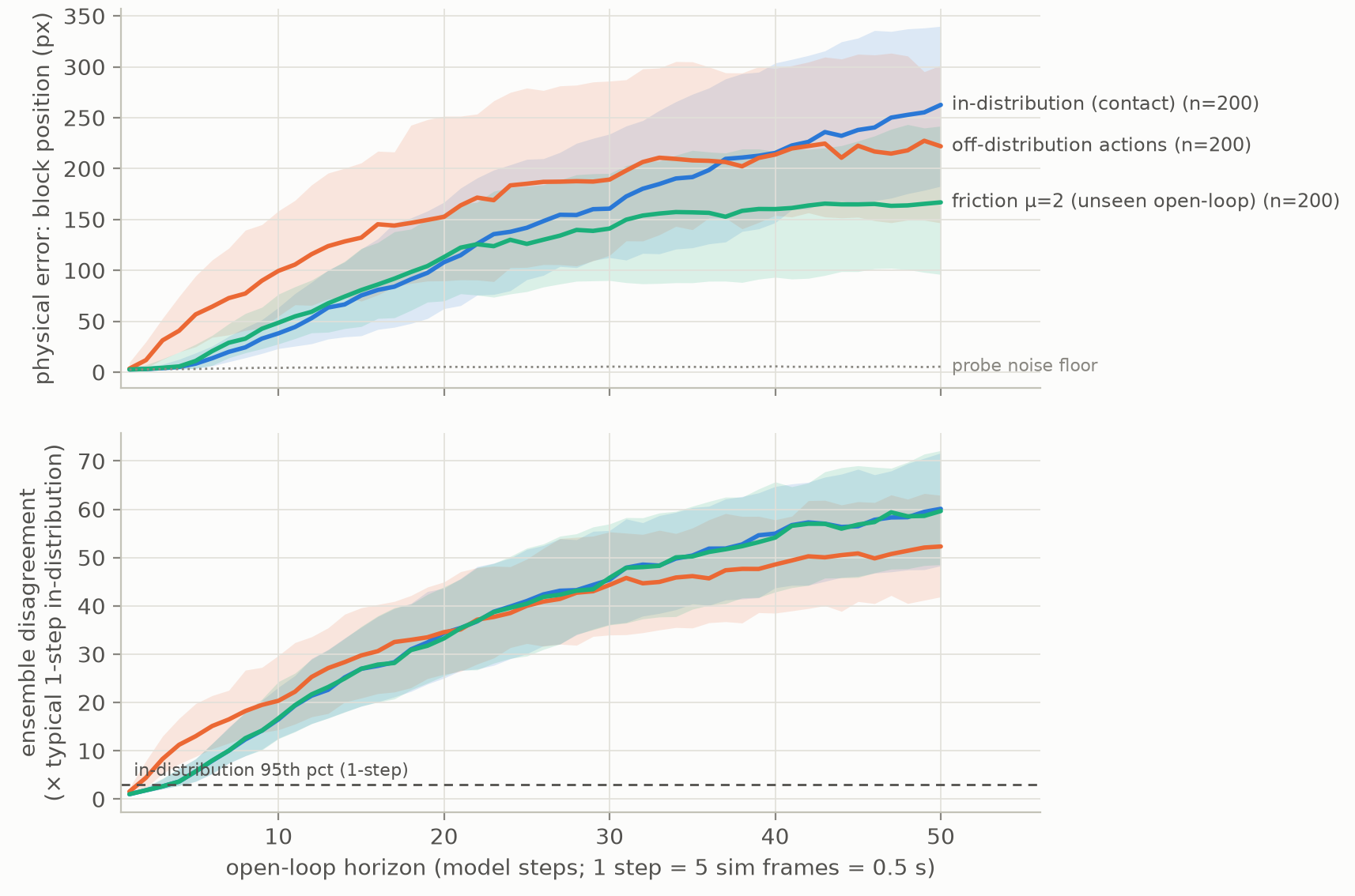}
\caption{Contact-rich rollouts. Off-distribution actions raise physical error
$2.6\times$ by step 10 (99 vs \SI{38}{px}) while disagreement rises only
$1.2\times$, and is \emph{lower} than in-distribution by step 50.}
\label{fig:b1c}
\end{figure}

\begin{figure}[t]
\centering
\includegraphics[width=0.78\textwidth]{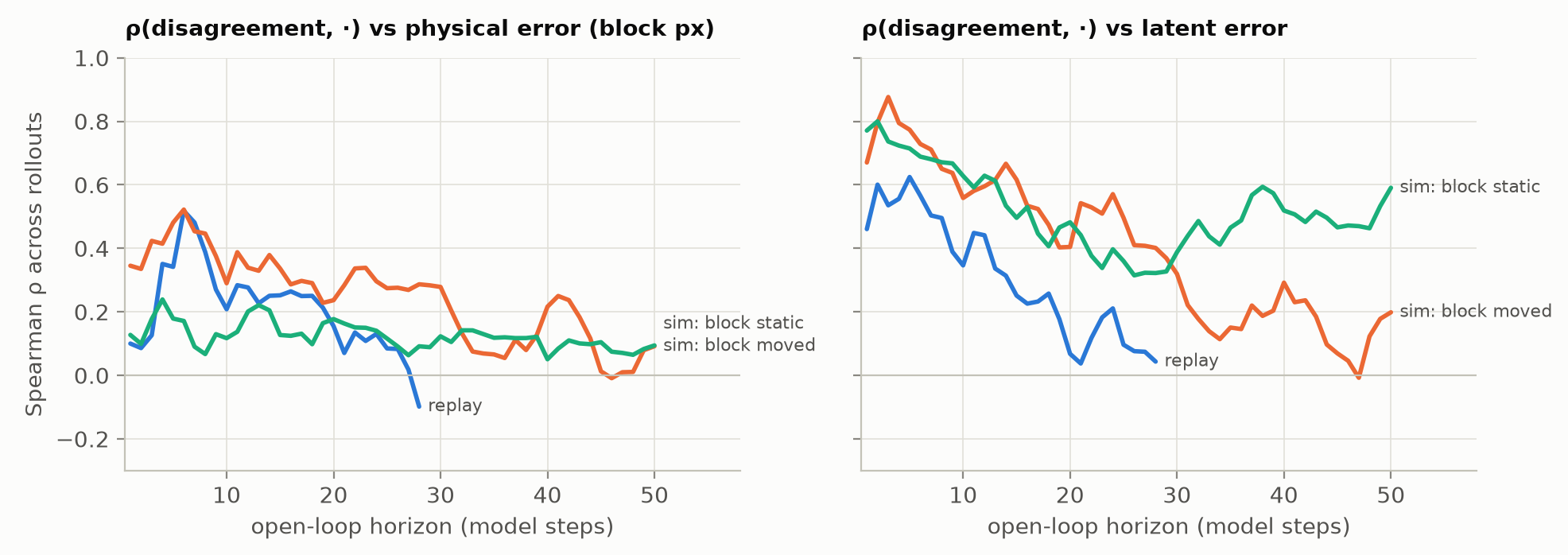}
\caption{Disagreement ranks latent error far better than physical error.}
\label{fig:b2}
\end{figure}

\begin{figure}[t]
\centering
\includegraphics[width=0.78\textwidth]{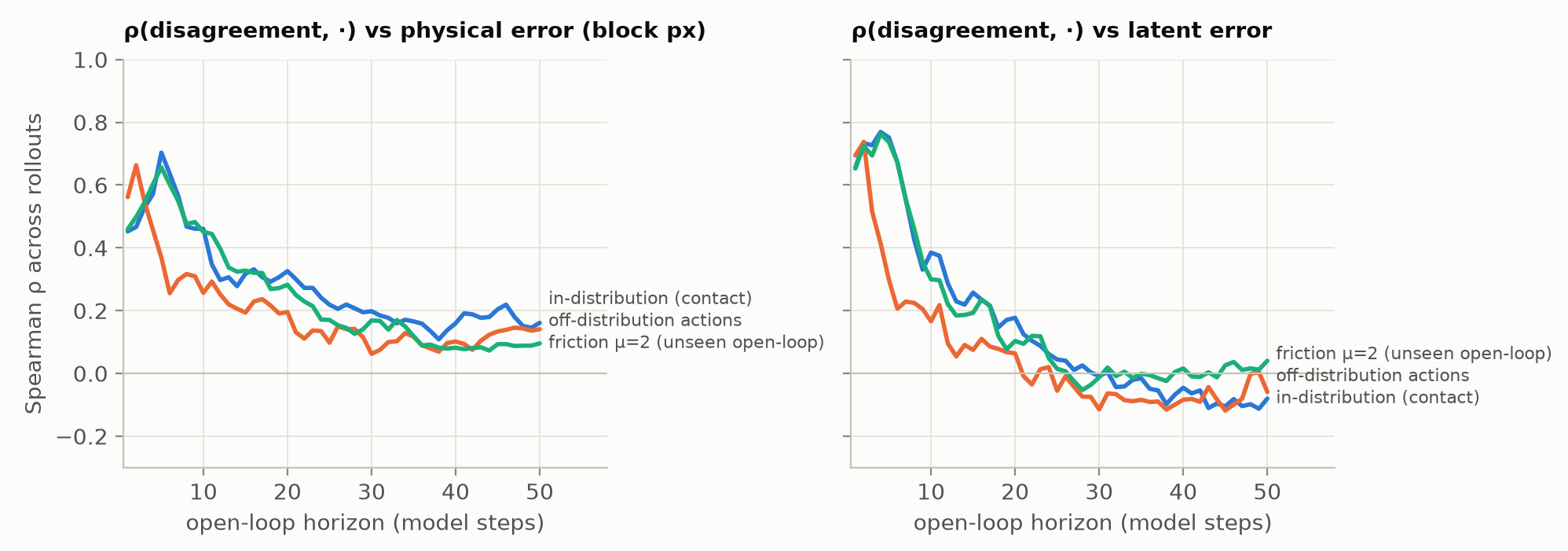}
\caption{Contact-rich conditions. Rank correlation with physical error decays to
\numrange{0.1}{0.2} by step 50 in all three conditions.}
\label{fig:b2c}
\end{figure}

\begin{figure}[t]
\centering
\includegraphics[width=0.62\textwidth]{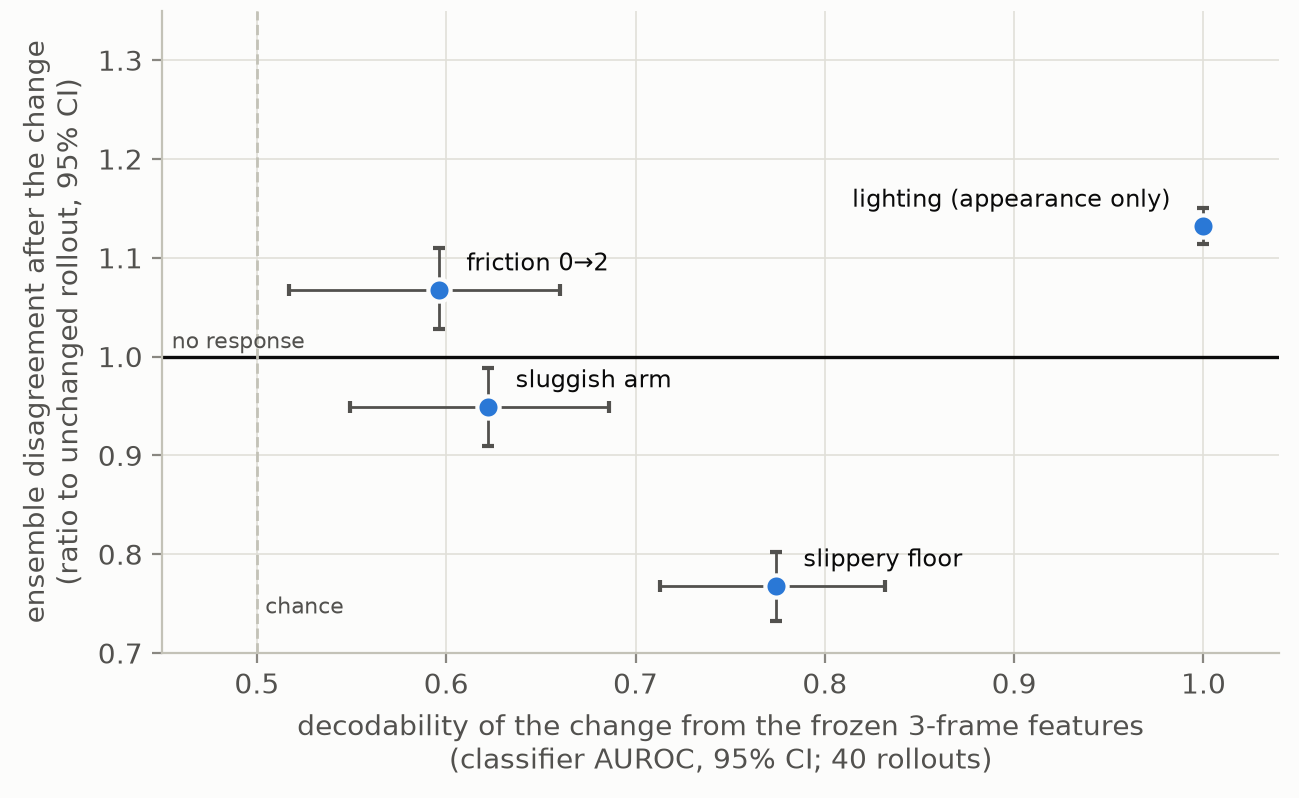}
\caption{How available a change is in the frozen features does not predict
whether disagreement responds to it. The appearance-only change is perfectly
decodable and produces the largest rise; the slippery floor is clearly
decodable and produces a fall.}
\label{fig:b9}
\end{figure}

\begin{figure}[t]
\centering
\includegraphics[width=\textwidth]{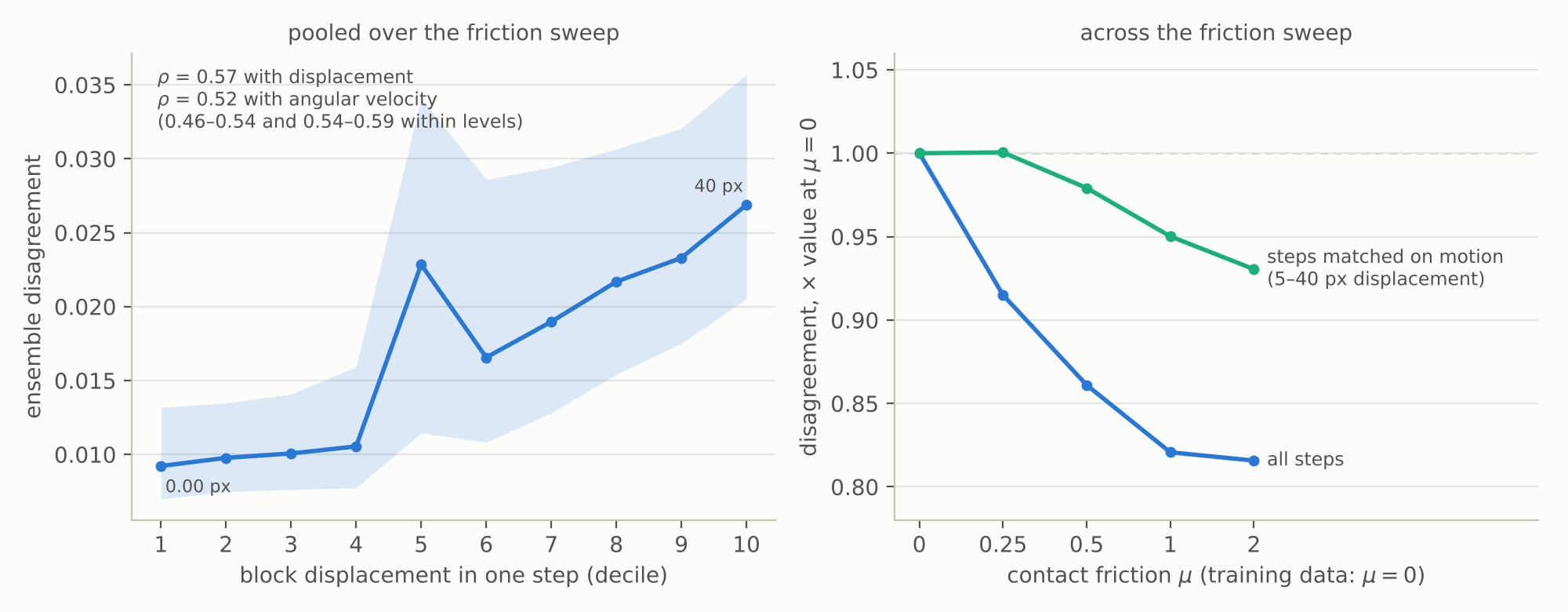}
\caption{What disagreement follows. Left: it rises with how far the block moved
in a step, from the lowest displacement decile to the highest, and the
correlation holds at every friction level separately. Right: across the
friction sweep it falls to $0.82\times$ its baseline, but among steps where the
block moved a comparable distance it is nearly flat ($\times 0.93$ at
$\mu = 2$). Most of the decline is the block moving less.}
\label{fig:b5}
\end{figure}

\begin{figure}[t]
\centering
\includegraphics[width=\textwidth]{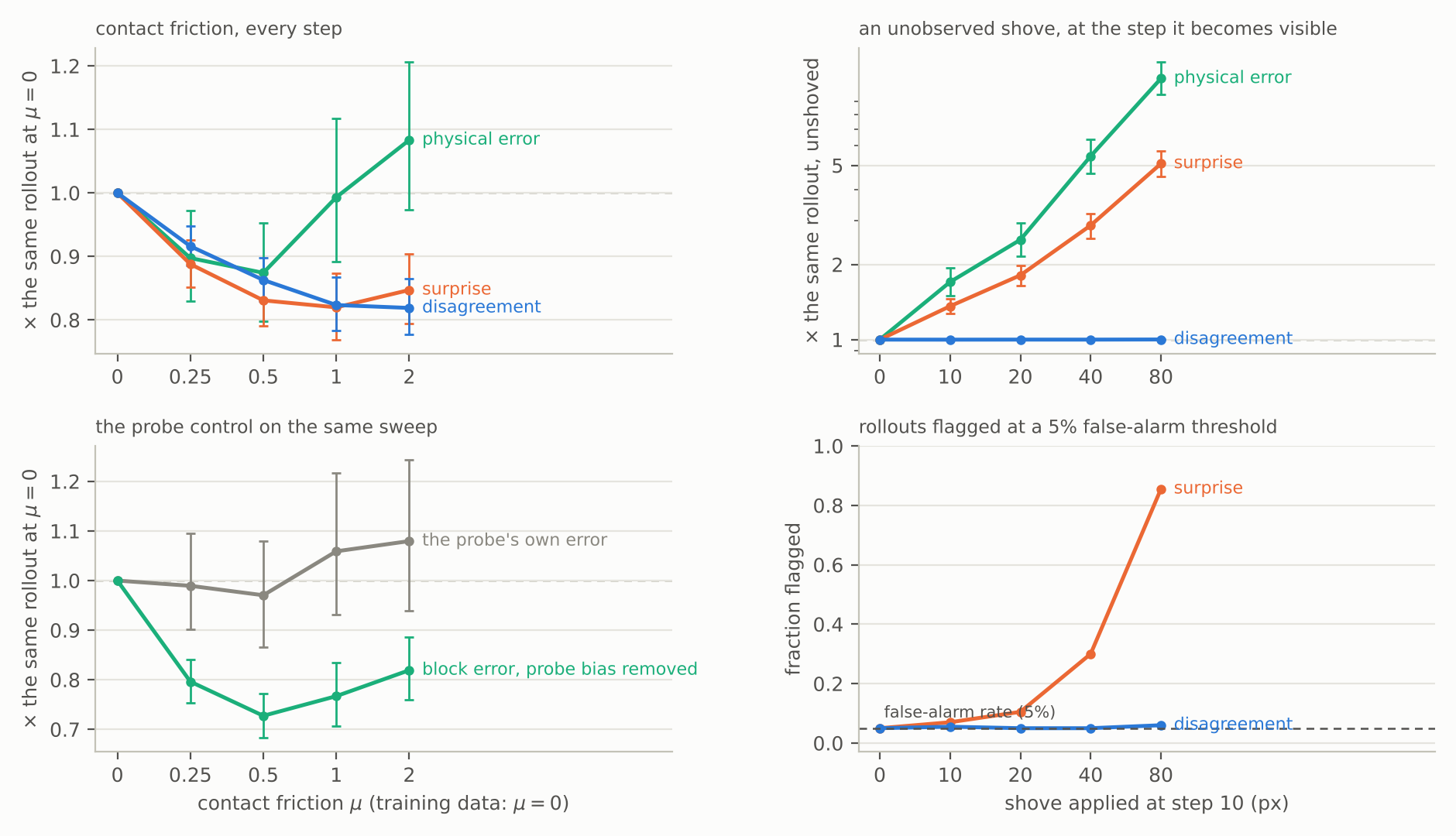}
\caption{The two severity sweeps, monitoring mode, 200 paired rollouts, every
quantity a geometric mean of per-rollout ratios against the same rollout
without the change (\S\ref{sec:setup}); bars are cluster bootstraps over
rollouts. \textbf{Left, contact friction.} The apparent rise in block error
($\times 1.08$ at $\mu = 2$) is the probe degrading over the sweep: its own
error on true frames rises by the same factor, and with that bias removed block
error \emph{falls} to $\times 0.82$. Disagreement also falls, to
$\times 0.82$, which Fig.~\ref{fig:b5} attributes to the block moving less.
Neither half of the apparent dissociation survives its control, so we report
this sweep as an explained negative rather than as evidence.
\textbf{Right, an unobserved shove.} Scored at the step the shove enters the target but not yet the model's input, physical error reaches $\times 11.2$ and surprise $\times 5.08$, while
disagreement is $\times 1.00$ at every severity, the shove not yet being in its
input. Surprise flags \SI{85}{\percent} of rollouts at \SI{80}{px} against a
\SI{5}{\percent} false-alarm rate; disagreement flags \SI{6}{\percent}.}
\label{fig:b3}
\end{figure}

Figures behind the supporting experiments of \S\ref{sec:trust} and
Appendix~\ref{sec:trustdetail}: member training and drift
(Figs.~\ref{fig:a1},~\ref{fig:a3}), the step counter that marks the horizon
confound (Fig.~\ref{fig:b2b}), open-loop error and ranking
(Figs.~\ref{fig:b1},~\ref{fig:b1c},~\ref{fig:b2},~\ref{fig:b2c}), decodability against response (Fig.~\ref{fig:b9}), what disagreement follows
(Fig.~\ref{fig:b5}), and the two severity sweeps (Fig.~\ref{fig:b3}).

\clearpage
\bibliographystyle{IEEEtran}
\bibliography{references}

\end{document}